\documentclass[journal]{IEEEtran}
\IEEEoverridecommandlockouts 
                                                          
\usepackage{hyperref, lipsum}
\usepackage{graphics} 
\usepackage{checkend}	
\usepackage{graphicx}	
\usepackage{amsmath,hyperref}
\usepackage[noabbrev]{cleveref}      
\usepackage[caption=false]{subfig}
\usepackage{epsfig} 
\usepackage{times} 
\usepackage{textcomp,gensymb}
\usepackage{mathtools}
\usepackage{makecell}
\usepackage{bm}
\usepackage[ruled]{algorithm2e}
\usepackage{siunitx}

\usepackage{xcolor}  
\usepackage[normalem]{ulem}

\begin{document}
\bstctlcite{IEEEexample:BSTcontrol}

\title{Hybrid Robot-assisted Frameworks for Endomicroscopy Scanning in Retinal Surgeries*}
\author{Zhaoshuo Li$^{\dagger,1}$, Mahya Shahbazi$^{\dagger,1}$, Niravkumar Patel$^{1}$, Eimear O' Sullivan$^{2}$, \\Haojie Zhang$^{2}$, Khushi Vyas$^{2}$, Preetham Chalasani$^{1}$, Anton Deguet$^{1}$, \\Peter L. Gehlbach$^{3}$, Iulian Iordachita$^{1}$, Guang-Zhong Yang$^{2}$, Russell H. Taylor$^{1}$%

    \thanks{\textdagger ~ Authors contributed equally to the work.}
    \thanks{* This work was funded in part by: NSF NRI Grants IIS-1327657, 1637789; Natural Sciences and Engineering Research Council of Canada (NSERC) Postdoctoral Fellowship \#516873; Johns Hopkins internal funds; Robotic Endobronchial Optical Tomography (REBOT) Grant EP/N019318/1; EP/P012779/1 Micro-robotics for Surgery; and NIH R01 Grant 1R01EB023943-01.}
    \thanks{$^{1}$Authors with the Laboratory for Computational Sensing and Robotics, Johns Hopkins University, Baltimore, Maryland 21218, USA. Email: \textit{\{zli122\}@jhu.edu.}}%
    \thanks{$^{2}$Authors with the Hamlyn Centre for Robotic Surgery, Imperial College London, SW7 2AZ, London, UK.}%
    \thanks{$^{3}$Author with the Johns Hopkins Wilmer Eye Institute, Johns Hopkins Hospital, 600 N. Wolfe Street, Maryland 21287, USA.}%

}

\IEEEtitleabstractindextext{%
\begin{abstract}
High-resolution real-time intraocular imaging of retina at the cellular level is very challenging due to the vulnerable and confined space within the eyeball as well as the limited availability of appropriate modalities. A probe-based confocal laser endomicroscopy (pCLE) system, can be a potential imaging modality for improved diagnosis. The ability to visualize the retina at the cellular level could provide information that may predict surgical outcomes. The adoption of intraocular pCLE scanning is currently limited due to the narrow field of view and the micron-scale range of focus. In the absence of motion compensation, physiological tremors of the surgeons’ hand and patient movements also contribute to the deterioration of the image quality.

Therefore, an image-based hybrid control strategy is proposed to mitigate the above challenges. The proposed hybrid control strategy enables a shared control of the pCLE probe between surgeons and robots to scan the retina precisely, with the absence of hand tremors and with the advantages of an image-based auto-focus algorithm that optimizes the quality of pCLE images. The hybrid control strategy is deployed on two frameworks - cooperative and teleoperated. Better image quality, smoother motion, and reduced workload are all achieved in a statistically significant manner with the hybrid control frameworks.
\end{abstract}}

\maketitle
\IEEEdisplaynontitleabstractindextext
\section{Introduction}
Among the various vision-threatening medical conditions, retinal detachment is one of the most common, occurring at a rate of about 1 in 10,000 per eye per year worldwide \cite{mitry2010epidemiology}. Mechanistically retinal detachment represents a separation of the neural retina from the necessary underlying supporting tissues, such as the retinal pigment epithelium and the underlying choroidal blood vessels, which collectively provide multiple types of support required for retinal survival. If not treated promptly, retinal detachment can result in a permanent loss of vision. One possible regenerative therapeutics treatment to the injured but not dead retina is to deliver a neuroprotective agent, such as stem cells, to targeted retina cells. However, this requires finding the viable retina cells identified by using cellular-level information \cite{pardue2018neuroprotective}.

A recent promising technique for \textit{in-vivo} characterization and real-time visualization of biological tissues at the cellular level is probe-based confocal laser endomicroscopy (pCLE). pCLE is an optical visualization technique that can facilitate cellular-level imaging of biological tissues in confined spaces. The effectiveness of pCLE has previously been demonstrated using real-time visualization of the gastrointestinal tract \cite{meining2007confocal}, thyroid gland \cite{wang2018robotic}, \textcolor{black}{breast \cite{zuo2015toward} and gastric \cite{ping2019modular} tissue}. The use of benchtop confocal microscopy systems for imaging externally mounted retinal tissue has been reported \cite{tan2012quantitative}\cite{ramos2013use}. However, the small footprint of a pCLE probe (typically on the order of 1 mm) makes pCLE a suitable technique for intraocular cellular-level visualization. To our knowledge, there is only one prior work for intraocular pCLE scanning of the retina (M. Paques, personal communication, March 22, 2020) using a contact-based CellVizio ProFlex probe (Mauna Kea Technologies). The work by M. Paques \textit{et. al.} only took one image via without attempting to scan a large area. The image was acquired manually without considering the safety of retina.

\begin{figure}[thpb]
    \centering
    \subfloat[]{\includegraphics[width=0.2\linewidth]{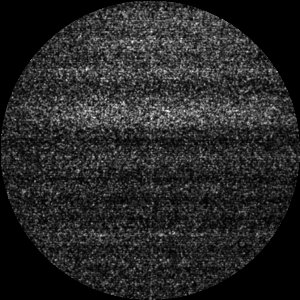}\label{fig:out_of_range}}   
    \hspace{0.01mm}
    \subfloat[]{\includegraphics[width=0.2\linewidth]{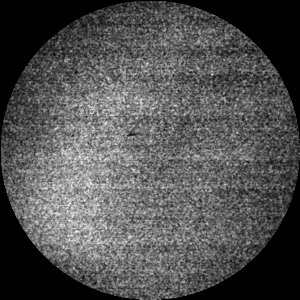}\label{fig:back_focus}}   
    \hspace{0.01mm}
    \subfloat[]{\includegraphics[width = 0.2\linewidth]{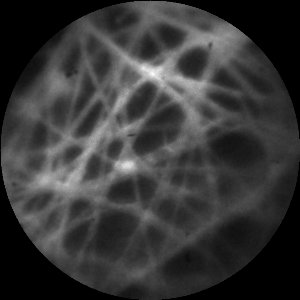}\label{fig:in_focus}}
    \hspace{0.01mm}
    \subfloat[]{\includegraphics[width=0.2\linewidth]{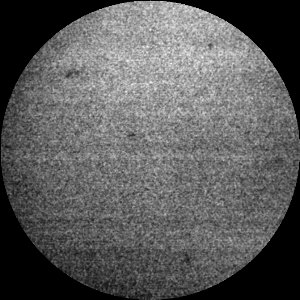}\label{fig:front_focus}}
    \captionsetup{width=\linewidth}
    \caption{Sample views of a non-contact pCLE system: (a) out-of-range, (b) back-focus, (c) in-focus, and (d) front-focus views, measured at a tissue distance of (a) 2.34~mm, (b) 1.16 mm, (c) 0.69 mm, and (d) fully in contact with the tissue, respectively.}
    \label{fig:probe_view}
\end{figure}

There are, however, several steep challenges to manual intraocular scanning of the retina using a pCLE probe for real-time \textit{in-vivo} diagnosis. First, the exceeding fragility and non-regenerative nature of retinal tissue do not allow any damage. It has been shown that forces as low as 7.5 mN can tear the retina \cite{gupta1999surgical}, thus necessitating a \textit{non-contact} scanning probe. Previous publications of contactless \textit{in vivo} cornea scanning \cite{pritchard2014non}\cite{pascolini2019non} promise the no-contact scanning working principles. Furthermore, due to the small size of pCLE fiber bundles, pCLE has a narrow field of view, usually on the sub-millimeter scale \cite{giataganas2015force}. Considering that the average diameter of the human eye is approximately 23.5 mm, one single pCLE image only covers a tiny portion of the scanning region. A solution to this is to use mosaicking algorithms. 
However, mosaicking algorithms require consistent high-quality images for feature matching and image stitching. The micron-scale focus range of a pCLE system, however, makes the manual acquisition of high-quality images extremely difficult; as the image quality drops considerably beyond the optimal focus range \cite{varghese2017framework}. For example, the non-contact pCLE system with a distal-focus lens in this study has an optimal focus distance of approximately 700 \micro m and a focus range of approximately 200 \micro m.  
\autoref{fig:probe_view} shows a comparison of our non-contact probe for four cases: 1) out-of-range, 2) back-focus, 3) in-focus, and 4) front-focus, measured at a probe-to-tissue distance of 2.34 mm, 1.16 mm, 0.69 mm, and fully in contact with the tissue, respectively. The physiological tremor of the human hand is on the order of magnitude of several hundreds of microns \cite{singhy2002physiological}\textcolor{black}{\cite{zhang2020hand}}, making it almost impossible to consistently maintain the pCLE probe in the optimal range required for high-quality image acquisition while assuring no or minimal contact.  
In addition to the above, patient movement, as well as the movement of the detached retina during the repair procedure, augment the complexities of manual scanning of the detached retinal tissue.

These challenges highlight the need for a robot-assisted manipulation of the pCLE probe for intraocular scanning of the retinal tissue. While robotic systems have been developed in the literature to mitigate the difficulties of manipulating pCLE probes, most are not yet suitable for intraocular use. In \cite{latt2011hand}, a hand-held device with a voice-coil actuator and a force sensor is presented to enable consistent probe-tissue contact. This device uses the measured forces to control the probe motion relative to the tissue. In \cite{erden2013conic}, a hollow tube is added surrounding the pCLE probe to provide friction-based adherence during contact with the tissue to enable a ``steady motion''. All the discussed techniques, however, are only compatible with \textit{contact-based} pCLE probes, which together with the large footprint of the devices, making them inappropriate for scanning the delicate retinal tissues inside of the confined space of the eyeball.

Robotic integration of the pCLE probe that does not require probe-tissue contact has also been achieved in \cite{zhang2017autonomous}, by presenting a system that tracks the tip position of the pCLE with an external tracker. However, in the present work, the scanning task was performed on a flat surface that presupposed the geometry and position of the tissue. Nonetheless, the curvature of the eyeball and the uneven surface of detached retina rapidly void these assumptions, rendering this system inapplicable to retinal surgery. 

\begin{figure}[!htpb]
    \centering
    \subfloat[]{\includegraphics[width=0.65\linewidth]{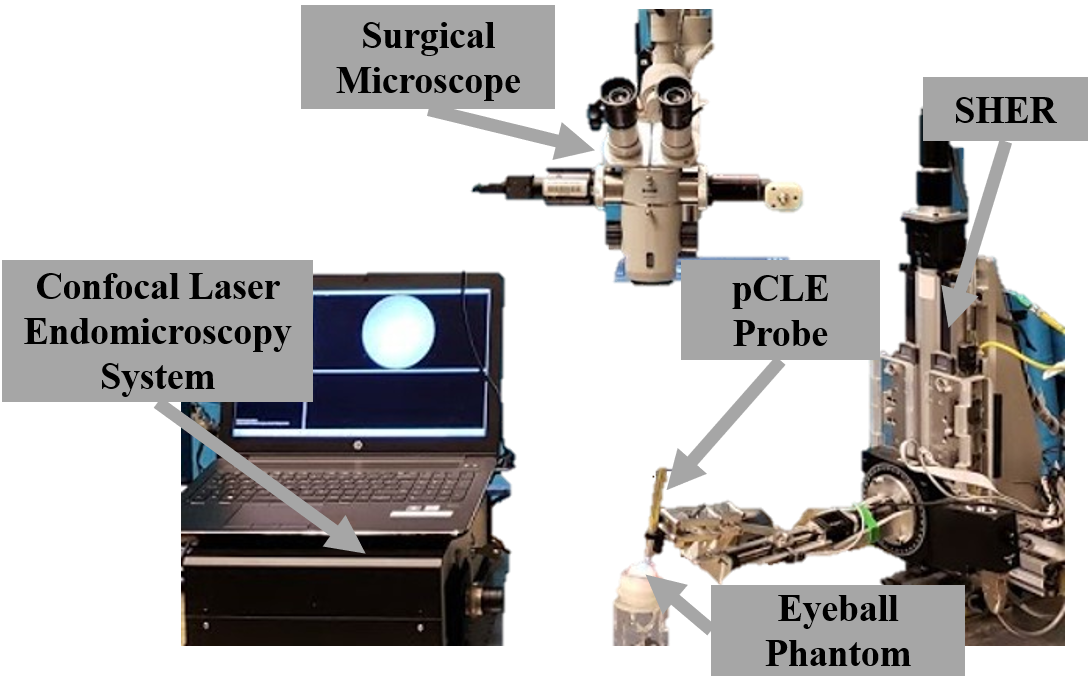}\label{fig:exp_setup_confocal}}
    \hspace{8pt}
    \subfloat[]{\includegraphics[trim={0 -1.5cm 0 0},clip,width=0.3\linewidth]{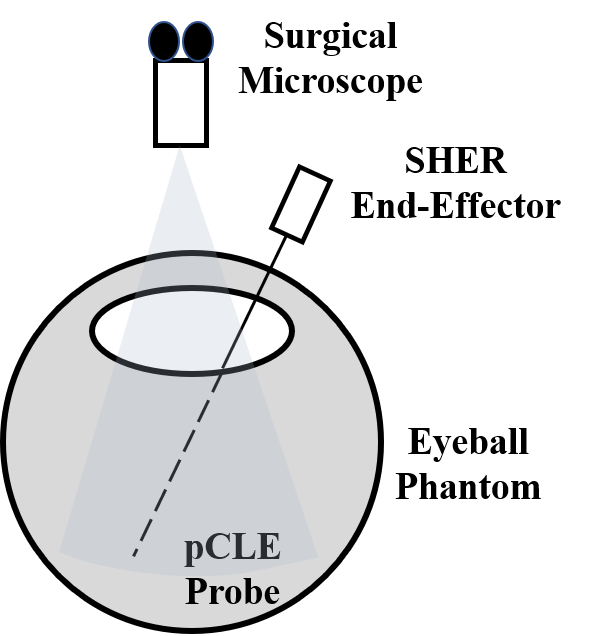}}
    
    \subfloat[]{\includegraphics[width=0.5\linewidth]{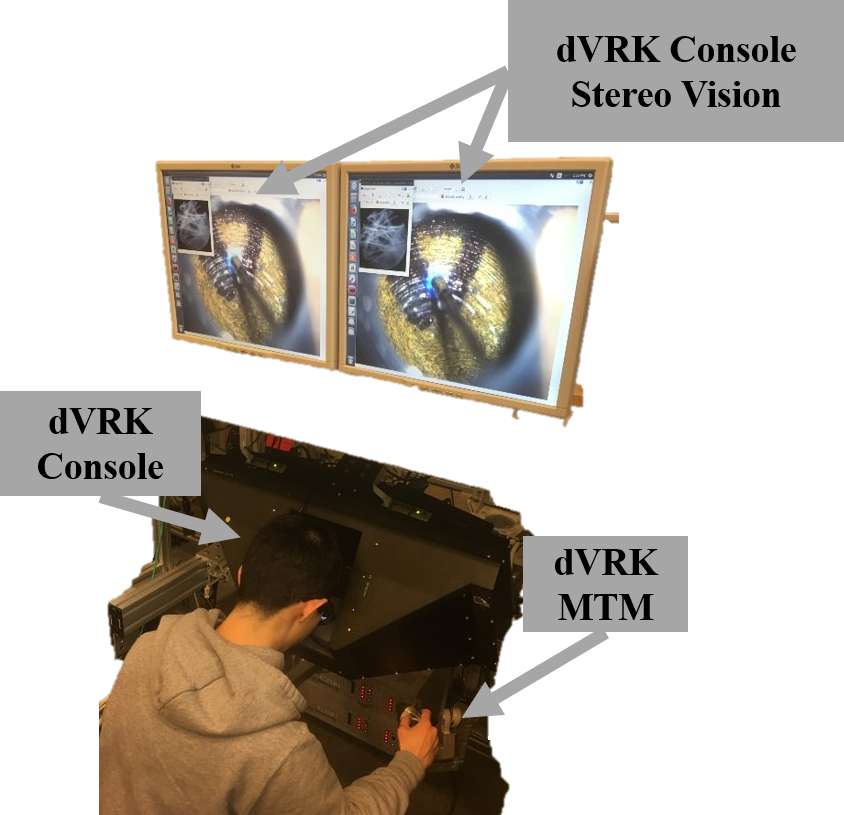}\label{fig:exp_setup_dvrk}}
    \hspace{10pt}
    \subfloat[]{\includegraphics[width=0.43\linewidth]{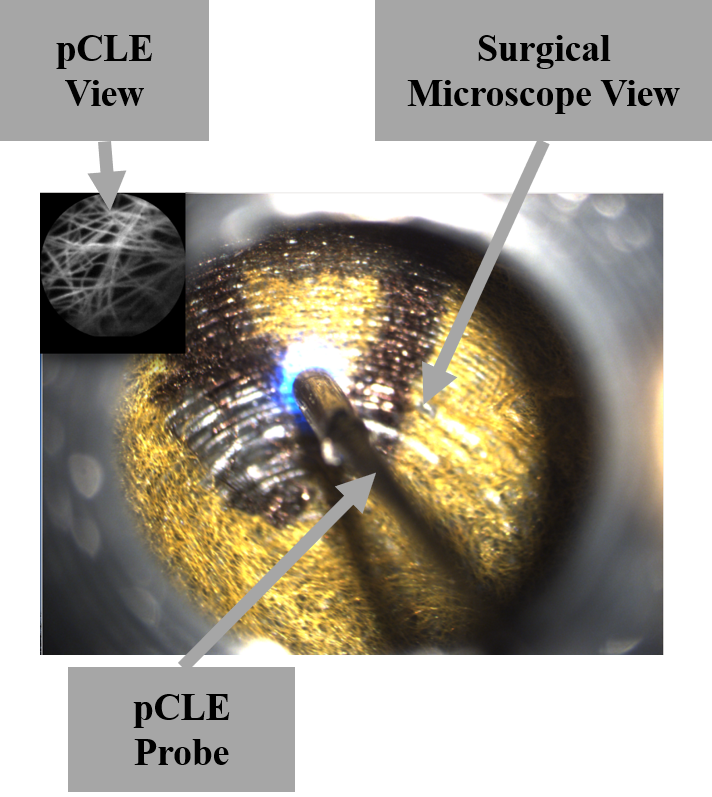}\label{fig:user_view}}
    
    \caption{The complete setup: (a) the SHER robot, the surgical microscope, the confocal laser endomicroscopy system, the pCLE probe, and an artificial eyeball phantom, (b) sketch of the setup, (c) the dVRK stereo vision, console and MTM, (d) the user's view, where the pCLE view is overlaid on the top left of the surgical microscope view looking through the eyeball opening, and the pCLE probe can be seen.}
    
    \label{fig:Experiment_Setup}
\end{figure}

Therefore, in this paper, a semi-autonomous hybrid approach is proposed for real-time pCLE scanning of the retina. Features of the proposed hybrid approach are briefly described:
\begin{itemize}
    \item a robot-assisted control that allows the surgeon to maneuver the probe with micron-level prevision laterally along the tissue surface freely, where the use of a robot eliminates the effect of hand tremor,
    \item an auto-focus algorithm based on the pCLE images that actively and optimally adjusts the probe-to-tissue distance for best image quality and enhanced safety.
\end{itemize}
The hybrid control strategy is deployed on two frameworks, cooperative and teleoperated, which have been shown to significantly improve the position precision in retinal surgery \cite{gijbels2014experimental}. Both implementations use the Steady-Hand Eye Robot (SHER, developed at LCSR, Johns Hopkins University
) end-effector to hold the customized non-contact fiber-bundle pCLE imaging probe. The pCLE probe is connected to a high-speed Line-Scan Confocal Laser Endomicroscopy system (developed at the Hamlyn Centre, Imperial College \cite{hughes2016line}). The teleoperated framework includes the da Vinci Research Kit (dVRK, developed at LCSR, Johns Hopkins University) 
to allow surgeons to remotely control the pCLE probe using the mater tool manipulator (MTM) and the stereo-vision console. A surgical microscope is used to visualize the surgical scene within the eyeball. \autoref{fig:Experiment_Setup} shows the complete setup.

The proposed hybrid cooperative and teleoperated frameworks have been validated through a series of experiments and a set of user studies, in which they are compared to manual operation, traditional cooperative and teleoperated control systems. We have discovered that the proposed hybrid frameworks result in statistically significant improvements in image quality, motion smoothness, and user workload.

\textbf{\textit{Contribution:}} To the best of our knowledge, this work is the first reported attempt to perform intraocular pCLE scanning of a large area of the retina with a distal-focus probe and with robot assistance. The introduction of an auto-focus, robot-assisted, non-contact pCLE scanning system will resolve previously discussed challenges and provide surface information of retina with enhanced safety and efficiency. The technical contribution reported can be summarized as follows:
\begin{itemize}
    \item A novel hybrid teleoperated control framework is proposed to enable significant motion scaling for improved precision of pCLE control, thus leading to consistently high-quality images. An improved hybrid cooperative framework extending the previously system in \cite{li2019novel} is also presented.
    \item The proposed hybrid controller includes a novel auto-focus algorithm to find the optimal position for best image quality while enhancing operation safety by avoiding probe-retina contact. A prior model of the retina curvature is included to speed up the algorithm and relax the assumption of scanning a flat surface within a static eye, which limited our previous work in \cite{li2019novel}. 
    \item Three experiment scenarios validate the challenge of manual scanning, the efficacy of the proposed auto-focus algorithm, and the enhanced smoothness of the scanning path and improved image quality using the hybrid framework over the traditional one.
    \item A user study involving 14 participants demonstrates the enhanced image-quality and reduced user workload in a statistically significant manner using the proposed hybrid system.
\end{itemize}

The remainder of the paper is organized as follows: \autoref{ssec:cooperative} explains the implementation of hybrid control strategy by using the hybrid cooperative framework; \autoref{ssec:teleoperated} presents the hybrid teleoperated framework; \autoref{sec:experiment} presents the three experiment evaluations; \autoref{sec:userstudy} presents the user study, results and discussion; \autoref{sec:conclusion} concludes the paper.

\section{Hybrid Cooperative Framework}
\label{ssec:cooperative}
The hybrid cooperative framework setup consists of the pCLE system and the SHER. The SHER is a cooperatively controlled robot with 5-Degrees-of-Freedom (DOF) developed for vitreoretinal surgery. It has a positioning resolution of 1 \micro m and a bidirectional repeatability of 3 \micro m. The pCLE system used for real-time image acquisition is a high-speed line-scanning fiber bundle endomicroscope, coupled to a customized probe (FIGH-30-800G, Fujikura Ltd.) of 30,000-core fiber bundle with distal micro-lens (SELFOC ILW-1.30, Go!Foton Corporation). The lens-probe distance is set such that the probe has an optimal focus distance of about 700 \micro m and a focus range of 200 \micro m.

\begin{figure*}[thpb]
    \centering
    \includegraphics[width=0.8\linewidth]{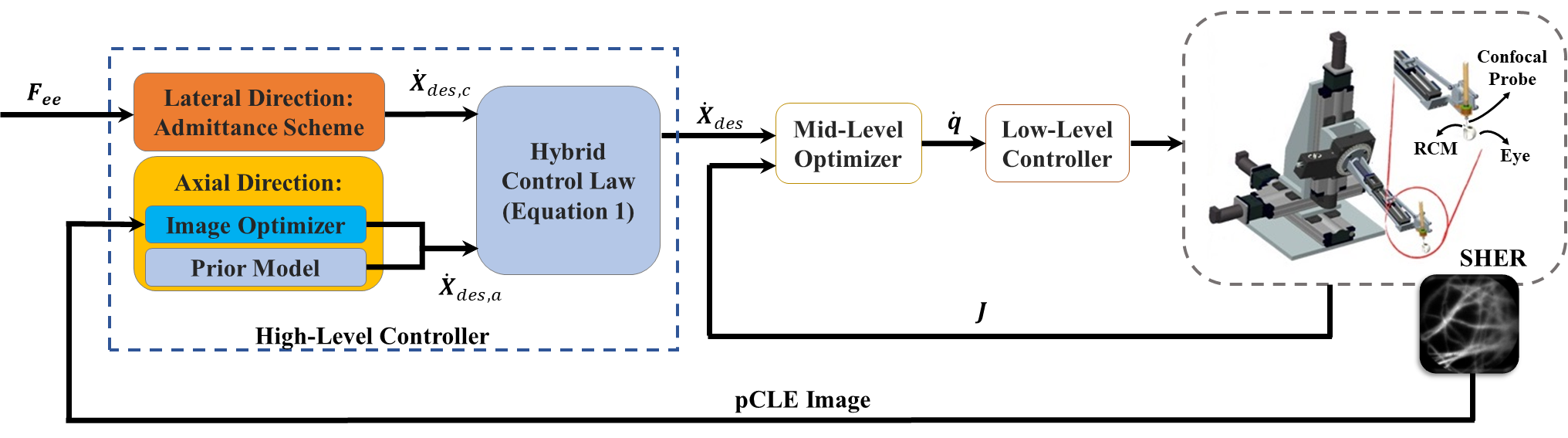}
    \caption{General schematic of the proposed hybrid cooperative control strategy. Hybrid Control Law is described in \autoref{eq:hybrid_motion_coop}. Lateral Direction control is described in \autoref{sssec:coop-lateral}. Axial Direction control is described in \autoref{sssec:coop-axial-grad} and \autoref{sssec:coop-axial-model}. Mid-Level Optimizer and Low-Level Controller are described in \autoref{sssec:coop-midlow}.}
    \label{fig:coop_block_diagram}
\end{figure*}

The hybrid cooperative framework consists of a high-level motion controller (which implements the semi-autonomous hybrid control strategy), a mid-level optimizer, and a low-level controller. \autoref{fig:coop_block_diagram} illustrates the block diagram of the overall closed-loop architecture.

\subsection{High-Level Controller}
The high-level controller enables dual functionality:
\begin{itemize}
    \item the control of the pCLE probe by surgeons with robot assistance in directions lateral to the scanning surface. Surgeons can maneuver the pCLE probe to scan regions of interest while the robot cancels out hand tremors.
    \item the autonomous control of the pCLE probe by the robot in the axial direction (along the normal direction of the tissue's surface) to actively control the probe-tissue distance for optimized image quality. This active image-based feedback can reduce the task complexity and the workload on surgeons while improving the image quality.
\end{itemize}
The four components of the high-level controller will be discussed separately.

\subsubsection{\textbf{Hybrid Control Law}}
\label{sssec:coop-hybrid}
The hybrid control law can be formulated as:
\begin{equation}
    \label{eq:hybrid_motion_coop}
    \dot{X}_{des} = {K}_c \dot{X}_{des,c} + {K}_a \dot{X}_{des,a}
\end{equation}
where subscripts $des$, $c$, and $a$ denote the desired hybrid motion, the cooperative motion by the surgeon (used for lateral control), and the auto-focus motion by the robot (used for axial control), respectively; $X$ denotes to the Cartesian position of the pCLE probe tip expressed in the base frame of the SHER. ${K}_c$ and ${K}_a$ denote the motion specification matrices \cite{khatib1987unified} that map a given motion to the lateral and axial directions of the retina surface respectively, and are defined as follows:
\begin{equation}
    K_c = \begin{bmatrix}
    R^{T}\Sigma_c R & 0_{3\times3}  \\
    0_{3\times3} & 1_{3\times3}  \\
    \end{bmatrix}, \Sigma_c = \begin{bmatrix}
        1 & 0 & 0 \\
        0 & 1 & 0 \\
        0 & 0 & 0
    \end{bmatrix}
\end{equation}
\begin{equation}
    K_a = \begin{bmatrix}
    R^{T}\Sigma_a R & 0_{3\times3}  \\
    0_{3\times3} & 0_{3\times3}  \\
    \end{bmatrix}, \Sigma_a = \begin{bmatrix}
        0 & 0 & 0 \\
        0 & 0 & 0 \\
        0 & 0 & 1
    \end{bmatrix}
\end{equation}
where ${R}$ denotes the orientation of the tissue normal expressed in the base frame of the SHER robot. 

\subsubsection{\textbf{Lateral Direction - Admittance Scheme}}
\label{sssec:coop-lateral}
The motion of the pCLE probe along the lateral direction is set to be controlled cooperatively \cite{taylor1999steady} by the SHER and the surgeon together to enable tremor-free scanning of regions of interest. Following \cite{taylor1999steady}, this is realized using an admittance control scheme, as follows:
\begin{align}
    \begin{split}
        \dot{X}_{des,ee} &= \alpha{F}_{ee}  \\
        \dot{X}_{des,c}&= Ad_{r,ee} \dot{X}_{des,ee}
    \end{split}
\end{align}
where subscripts $ee$ and $r$ indicate the end-effector frame and the base frame of the SHER, respectively; ${F}_{ee}$ denotes the interaction forces applied by the user's hand, expressed in the end-effector frame. ${F}_{ee}$ is measured by a 6-DOF force/torque sensor (ATI Nano 17, ATI Industrial Automation, Apex, NC, USA) attached on the end-effector; $\alpha$ is the admittance gain; ${Ad}_{r,ee}$ is the adjoint transformation matrix that maps the desired motion in the end-effector frame to the base frame of the SHER, and is calculated as 
\begin{equation}
  Ad_{r,ee} = \left[ \begin{array}{ccc}
{R}_{r,ee}  & skew({p}_{r,ee}){R}_{r,ee}  \\
0_{3\times3} & {R}_{r,ee} 
\end{array} 
\right ]  
\end{equation}
where ${R}_{r,ee}$ and ${p}_{r,ee}$ are the rotation matrix and translation vector of the end-effector frame expressed in the base frame of the SHER. Also, $skew({p}_{r,ee})$ denotes the skew-symmetric matrix associated with the vector ${p}_{r,ee}$. The resultant $\dot{X}_{des,c}$, along with the motion specification matrix $K_c$, specifies the lateral component of the desired motion given in \autoref{eq:hybrid_motion_coop}. 

\subsubsection{\textbf{Axial Direction - Image Optimizer}}
\label{sssec:coop-axial-grad}
The desired motion along the axial direction, $\dot{X}_{des,a}$, is defined such that the confocal image quality is optimized through an auto-focus control using a \textit{gradient-ascent search}. The autonomous control is performed in a sensorless manner, meaning that no additional sensing modality is required for depth/contact measurement. As discussed previously, the confined space within the eyeball and the fragility of the retinal tissue makes it very challenging, if not impossible, to add extra sensing modalities. The proposed sensorless and contactless control strategy relies on the image quality as an indirect measure of the probe-to-tissue distance, intending to maximize the quality autonomously.

Using image blur metrics as a depth-sensing modality has been previously discussed in \cite{varghese2017framework}. The control strategy presented therein, however, is only applicable to contact-based pCLE systems probes, which is not suitable for scanning the delicate retinal tissue. Besides, the control system presented in \cite{varghese2017framework} is dependent on the characteristics of the tissue and requires a pre-operative calibration phase. The calibration process necessitates pressing the contact-based probe onto the tissue and collecting a series of images from the pCLE system along with the force values applied to the tissue. However, this calibration process (which requires exertion and measurement of force values) is not
feasible for retinal scanning due to its fragility. 
Therefore, in this paper, a image-based auto-focus methodology is proposed for optimizing image sharpness and quality during \textit{non-contact} pCLE scanning without the necessity for any extra sensing modality. 

\begin{figure}[h]
    \centering
    \includegraphics[width=0.8\linewidth]{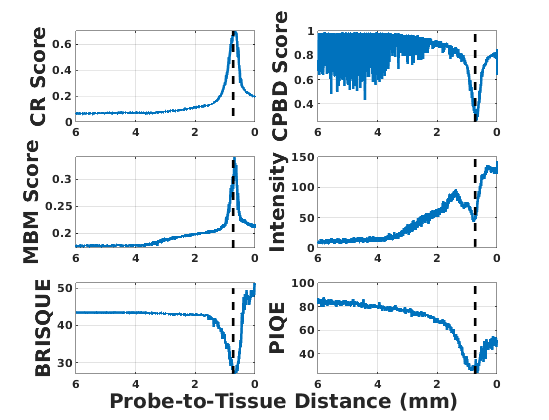}
    \caption{Evaluation of 6 image metrics with respect to (w.r.t) the probe-to-tissue distance (linear scale). The optimal view at a probe-to-tissue distance of 690 \micro m is indicated by the vertical dashed line.}
    \label{fig:plot_image_features}
\end{figure}

\begin{figure}[h]
    \centering
    \includegraphics[trim={8cm 1.5cm 6cm 3cm},clip,width=0.6\linewidth]{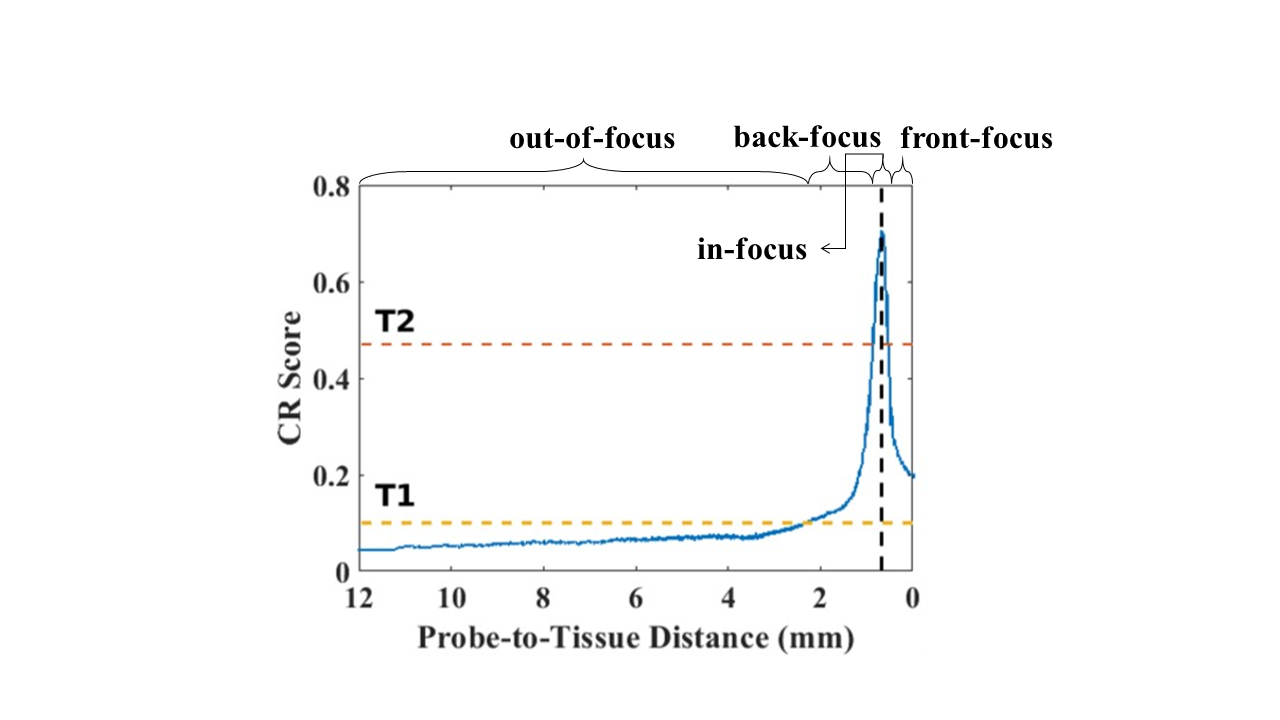}
    \caption{CR score w.r.t the probe-to-tissue distance, with the two horizontal lines indicating two thresholds $T_1$ and $T_2$, the vertical line indicating the optimal probe-to-tissue distance of approximately 690 \micro m. The four focus regions (out-of-focus, back-focus, in-focus, and front-focus) are labeled respectively.}
    \label{fig:plot_cr}
\end{figure}

For this purpose, the effectiveness of several blur metrics was evaluated to use in real-time control of the robot. The selected metrics are: Cr\'et\'e-Roffet (CR) \cite{crete2007blur}, Marziliano Blurring Metrix (MBM) \cite{marziliano2002no}, Cumulative Probability of Blur Detection (CPBD) \cite{narvekar2009no}, Blind/Referenceless Image Spatial QUality Evaluator (BRISQUE) \cite{mittal2012no} and Perception-based Image Quality Evaluator (PIQE) \cite{venkatanath2015blind}, which are all standard no-reference image metrics for quality assessment. Besides, the image intensity is calculated since the pCLE image will appear dark when back-focus and bright when front-focus. An experiment was conducted by commanding the SHER to move from a far distance to almost touching a scanned tissue surface. The six metrics were calculated for the pCLE images during the experiment, as shown in \autoref{fig:plot_image_features}. The optimal image view in this experiment was achieved around a probe-to-tissue distance of 690 \micro m. A consistent pattern was observed for the six metrics around the optimal view (\textit{i.e}., maximized value for MBM and CR scores, and minimized value for intensity, CPBD, BRISQUE, and PIQE). Among these six metrics, the lowest level of noise and the highest signal-to-noise ratio belongs to CR, which was eventually chosen to be incorporated into our auto-focus control strategy. CR is a no-reference blur metric and can evaluate both motion and focal blurs. It has a low implementation cost and high robustness to the presence of noise in the image. All of the above attributes make CR an efficient and effective metric for real-time image-based control. Moreover, since pCLE images exhibit blur patterns that CR can capture (as shown in \autoref{fig:probe_view} and \autoref{fig:plot_cr}), CR becomes a preferred metric to evaluate the clarity and indicate the clinical diagnosis value of a given pCLE image.

The working principle of the CR score is based on the fact that blurring an already blurred image does not result in significant changes in intensities of neighboring pixels, unlike blurring a sharp image. The CR score for an image $I$ of size $M \times N$ can be obtained by first calculating the accumulated absolute backward neighborhood differences in the image along the \textit{x} and \textit{y} axes, denoted as $d_{I,x}$ and $d_{I,y}$:

\begin{equation}
    \label{eqn: neighbour-diff}
    \begin{aligned}    
        d_{I,x} = \sum_{i,j=1}^{M,N} |I_{i,j}-I_{i-1,j}| \\
        d_{I,y} = \sum_{i,j=1}^{M,N} |I_{i,j}-I_{i,j-1}|
    \end{aligned}
\end{equation}

Then, the input image $I$ is convolved with a low-pass filter to obtain a blurred image $B$, also of size $M \times N$. Afterwards, the changes of neighborhood differences due to the low-pass filtration, denoted as $d_{IB,x}$ and $d_{IB,y}$, are calculated along \textit{x} and \textit{y} axes with a minimum value of 0: 
\begin{equation}
    \label{eqn: neighbour-diff-change}
    \begin{aligned}
    d_{IB,x} &= \sum_{i,j=1}^{M,N} max(0,|I_{i,j}-I_{i-1,j}| - |B_{i,j}-B_{i-1,j}|)\\
    d_{IB,y} &= \sum_{i,j=1}^{M,N} max(0,|I_{i,j}-I_{i,j-1}| - |B_{i,j}-B_{i,j-1}|)\\
    \end{aligned}
\end{equation}
The accumulated changes $d_{IB,x}$ and $d_{IB,y}$ are then normalized using the values obtained in \autoref{eqn: neighbour-diff}, indicating the level of blur present in the input image as
\begin{equation}
    \begin{aligned}
    blur_{x} = \frac{d_{I,x}-d_{IB,x}}{d_{I,x}} \\
    blur_{y} = \frac{d_{I,y}-d_{IB,y}}{d_{I,y}}
    \end{aligned}
\end{equation}
The quality, $q$, of the image $I$, \textit{i.e.} the amount of high-frequency content, is then calculated as the maximum blur level along the \textit{x} and \textit{y} axes as: 
\begin{equation}
    q_I = 1 - max(blur_{x},blur_{y})
\end{equation}
\autoref{fig:plot_cr} shows an example of the CR score w.r.t the probe-to-tissue distance acquired from the previous experiment. The CR metric has minimal variation and the lowest value when the pCLE probe is out-of-focus (illustrated in \autoref{fig:out_of_range}); the maximum value (i.e. the image quality is maximized) when the probe-to-tissue distance is optimized and the pCLE probe is in-focus (illustrated in \autoref{fig:in_focus}); and the CR metric drops sharply when the probe is either back-focus (\autoref{fig:back_focus}) or front-focus (\autoref{fig:front_focus}).
Accordingly, the autonomous probe-to-tissue adjustment algorithm (\autoref{alg:image_feedback}) is designed based on a stochastic gradient-ascent approach to maximize the image quality.

An interesting observation in \autoref{fig:plot_cr} is that the metric has an almost symmetric pattern around the optimal in-focus probe-to-tissue distance, while having an asymmetric pattern in the out-of-range region. This asymmetry is used in our proposed framework to identify probe-to-tissue distances that are too far, and to avoid the vanishing gradient problem that may occur. For this purpose, a pre-defined threshold $T_1$ is chosen, below which the pCLE probe is considered to be out-of-focus. When the probe-to-tissue distance is large and the pCLE probe is far from the retina (e.g., when the pCLE probe is first inserted into the eyeball), the CR score drops below $T_1$. The surgeon can obtain full control of the pCLE probe in both the lateral and axial direction, and all autonomous movement is disabled. It should be noted that this transfer of control only happens when the probe is sufficiently far from the retina and in the out-of-range region. The robot takes over the axial control as soon as the probe approaches the retina and enters the back-focus region. This ensures the safety of the patient as there is no longer the risk of the surgeon suddenly touching or puncturing the retina tissue.

Due to the symmetry around the peak of the CR score (where the image quality is optimal and probe is in-focus), the history of the axial movement of the probe relative to the tissue is used to determine the gradient of the CR score w.r.t the probe-to-tissue distance, such that the direction to increase the CR score is found. If the relative motion of the probe has the same sign as the variation of the image score between the current state and the previous state, the robot is commanded by the algorithm to continue moving in the same direction, and otherwise, to reverse the direction of motion. For example, in the back-focus region, the probe will move closer to the tissue since the gradient is positive; in the front-focus region, the probe will move further from the tissue since the gradient is negative. Therefore, the relative desired movement $\Delta {X}_{des,a}$ is calculated based on the product of a gain value $g$ (to convert the image score to a distance value) and $1-q_I$. The term $1-q_I$ is used as an adaption factor of the step size, so that the motion of the probe is slowed down further as it gets closer to the optimal distance. The direction of the desired relative motion $\Delta {X}_{des,a}$ is determined by $SIGN(\Delta {X}_{probe} \Delta q ) \cdot SIGN(\Delta {X}_{probe})$, where $SIGN(\cdot)$ is the sign (positive or negative) of the input $(\cdot)$. 

In \autoref{alg:image_feedback}, a threshold $T_2$ is chosen as a termination condition, indicating that the optimal probe-to-tissue distance and, thereby, the in-focus condition is met. When an image quality higher than $T_2$ is achieved, the axial motion of zero is sent to the robot. Therefore, the axial motion of the robot is zero unless the score drops beyond the threshold $T_2$. This could happen with the patient motion or a sudden curvature change on the surface of the retina. In this scenario, the robot does not have a previous state of motion to rely on to identify the desired direction of motion. Therefore, it will perform an exploration step to specify the appropriate direction of motion. As a safety consideration, the exploration step is always away from the retina to ensure no contact between the pCLE probe and the retina.

\begin{figure}[thpb]
    \centering
    \subfloat[]{\includegraphics[width=0.2\linewidth]{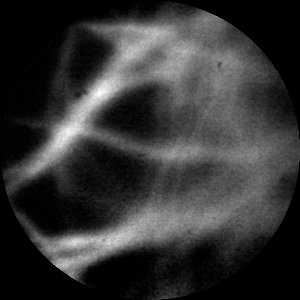}\label{fig:blur_close}}   
    \hspace{0.01mm}
    \subfloat[]{\includegraphics[width=0.2\linewidth]{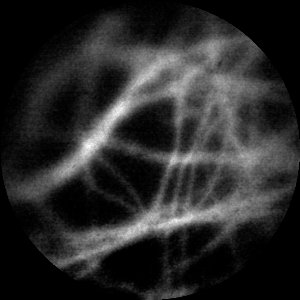}\label{fig:optimal}}   
    \hspace{0.01mm}
    \subfloat[]{\includegraphics[width = 0.2\linewidth]{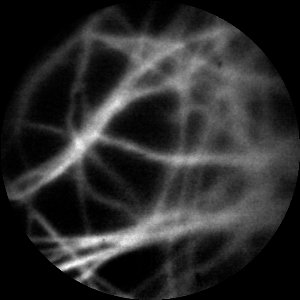}\label{fig:blur_far}}
    \captionsetup{width=\linewidth}
    \caption{Sample probe views of different CR score, (a) CR=0.45 at 0.51 mm, (b) CR=0.47 at 0.62 mm, and (c) CR=0.61 at 0.69 mm.}
    \label{fig:threshold}
\end{figure}

In this framework, a CR score of $0.47$, where the image view appears in-focus visually, is chosen for $T_2$. A CR score of $0.10$ is chosen for $T_1$, where the pCLE probe is out-of-range. A comparison of three images collected at CR values of $0.45, 0.47$ and $0.61$ is shown in \autoref{fig:threshold}. It should be mentioned that, since the proposed approach relies only on the gradient of the CR score, it is not sensitive to the exact shape of CR w.r.t the probe-to-tissue distance. Also, only knowing a rough estimation of the two threshold values would suffice for the algorithm, which can be specified pre-operatively. 

The desired axial velocity, ${\dot{X}}_{des,a}$ used in \autoref{eq:hybrid_motion_coop} is calculated based on the control-loop sampling time ${\Delta t}$ and the desired axial displacement $\Delta {X}_{des,a}$, as $\dot{X}_{des,a}=\frac{\Delta X_{des,a}}{\Delta t}$.

\begin{algorithm}[ht]
\SetAlgoLined
\KwIn{Image ($I$), Previous CR score ($q_{prev}$), Current/Previous probe position ($X_{probe,curr}, X_{probe,prev})$}
\KwOut{Desired Movement ($\Delta X_{des,a}$)}
 $q_I = CR(I)$\;
 $\Delta q = q_I - q_{prev}$ \;
 $\Delta X_{probe} = K_a (X_{probe,curr} - X_{probe,prev})$ \;
 \uIf{$q_I < T_1$}{
  \KwRet{$\Delta X_{des,a} = {\dot{X}}_{des,c} \Delta t$}\;
 }
 \uElseIf{$T_1 < q_I < T_2$}{
  \uIf{$ \Delta X_{probe} <$ ROBOT RESOLUTION $\And \Delta q < 0$ }
  {$\Delta X_{des,a} = g(1 - q_I)$\;} 
  \uElseIf{$\Delta X_{probe} <$ ROBOT RESOLUTION $\And \Delta q > 0$}
  {$\Delta X_{des,a} = \Delta X_{prev}$ \;}
  \Else{$\Delta X_{des,a} = g$SIGN$(\Delta X_{probe} \Delta q )$ SIGN$(\Delta X_{probe}) (1 - q_I)$\;}
 }
 \Else{
  $\Delta X_{des,a} = 0$ \;
 }
 \caption{Axial Control Algorithm, Image Optimizer}
 \label{alg:image_feedback}
\end{algorithm}

\subsubsection{\textbf{Axial Direction - Prior Model}}
\label{sssec:coop-axial-model}
\begin{figure}[h]
    \centering
    \subfloat[]{\includegraphics[trim={13cm 3.5cm 8cm 3cm},clip,width=0.4\linewidth]{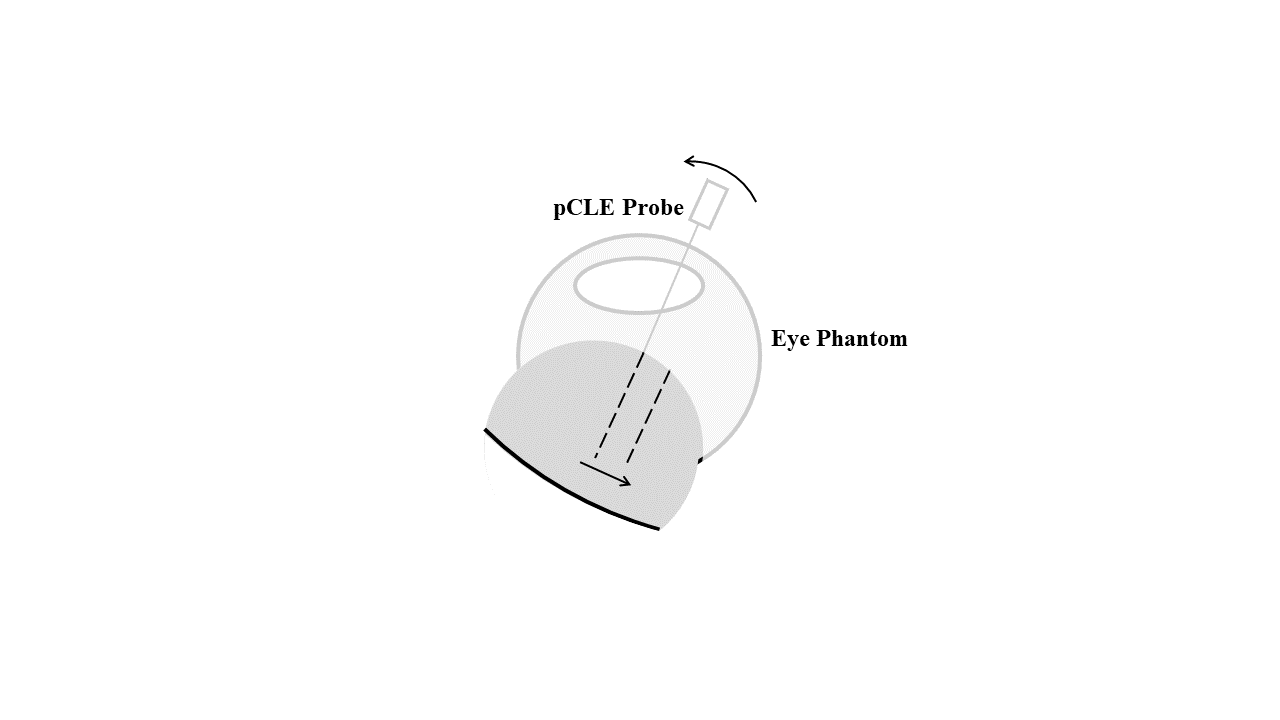}\label{fig:curved_surface}}
    \hspace{1pt}
    \subfloat[]{\includegraphics[width=0.45\linewidth]{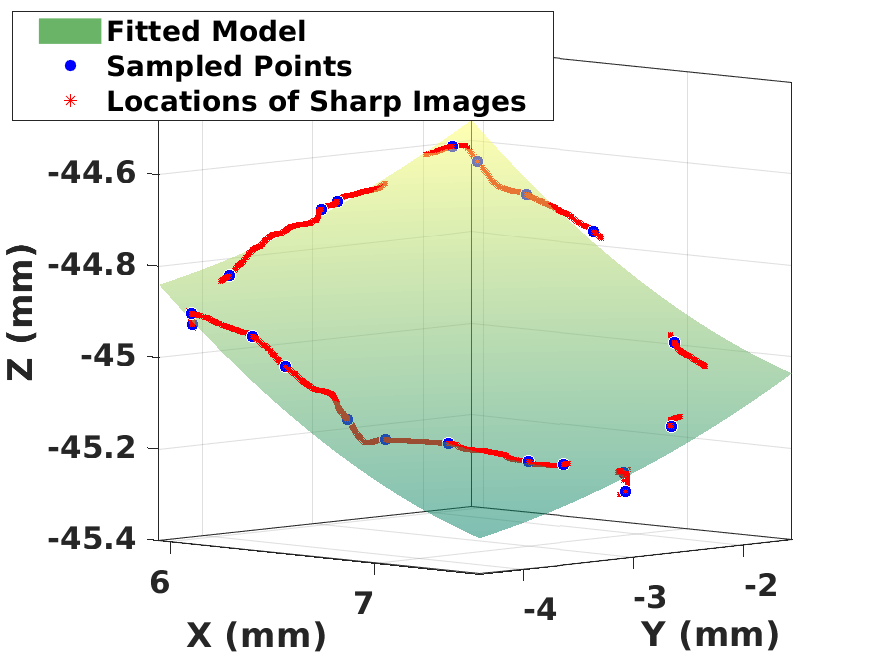}\label{fig:model_fitting}}
    
    \subfloat[]{\includegraphics[width=0.45\linewidth]{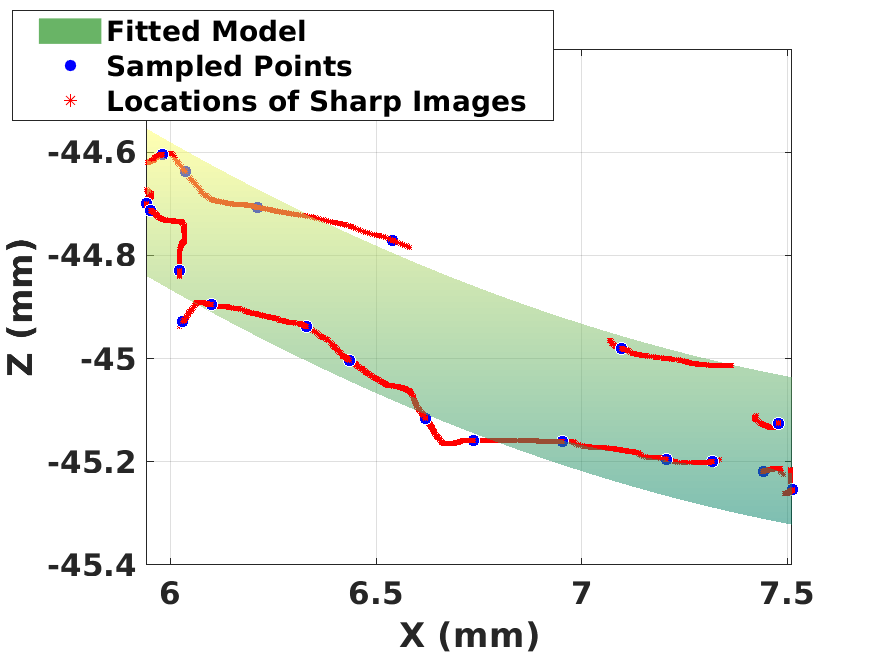}\label{fig:model_fitting_2}}
    \subfloat[]{\includegraphics[width=0.45\linewidth]{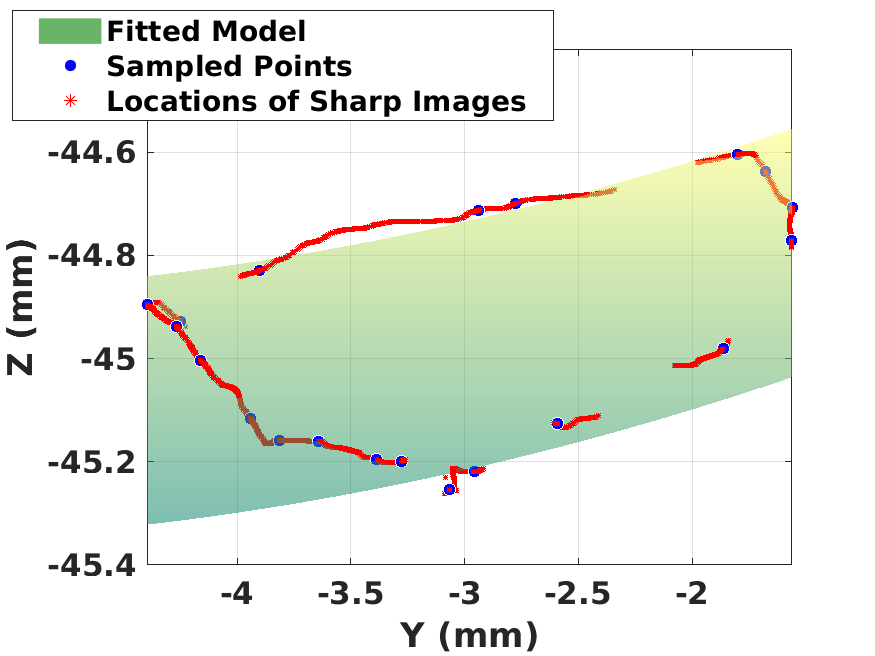}\label{fig:model_fitting_3}}
    \caption{(a) Illustration of the failure case for \autoref{alg:image_feedback}. Assuming the pCLE probe is currently in-focus. As the probe moves in the direction of the arrow, the image score will decrease due to the larger probe-to-tissue distance, without the probe movement in the axial direction. The safety feature implemented will move the probe away from the tissue. (b)(c)(d) Example of the fitted model, where (c) is the projected view of XZ plane and (d) is the projected view of YZ plane; Green surface: fitted local polynomial model of the retina. Red points: scanning path during the registration process. Blue points: sampled points where images are in-focus.}
\end{figure}

For safety purposes, \autoref{alg:image_feedback} is designed such that the exploration step is always away from the retina. When scanning a curved-surface, this retracted motion may move the probe to a gradient-diminishing region as shown in \autoref{fig:curved_surface} before the system captures the right direction of motion again. This can elongate the process of specifying a dominant gradient for correcting motion direction and, thereby, lead to poor user experience. 

To address this, a prior model of the retina is integrated into the algorithm to provide the algorithm with a suitable initialization state. To obtain the retina curvature model, the perimeter of the area of interest is scanned once. During the scanning, both the CR score and the probe position are recorded. Based on the CR score threshold $T_2$, the image qualities at different positions are classified as either in-focus or not. By spatially sampling 20 of the in-focus positions, a second-order parabola is fitted. The resulted model is denoted as $M(\cdot)$, where the input $(\cdot)$ is the lateral probe position $K_c X_{probe,curr}$, and the output is the desired axial position of the probe, both expressed in the base frame of the SHER. \autoref{fig:model_fitting} shows an example of prior model.

The prior model will not be accurate enough at the micron-scale to keep the probe focused directly, but it can augment the previously mentioned gradient-ascent approach (\autoref{alg:image_feedback}) to fine-tune the image quality. Thus, \autoref{alg:model_image} is proposed to take both the prior model and the gradient-ascent image optimizer into account. The flag $F_M$ indicates if the in-focus axial position has been reached according to the prior model. When the flag $F_M$ is set to true and the image quality is still far from desirable, the gradient-ascent approach will be then activated to further fine-tune and improve the image quality. If the current position of the probe tip $ X_{probe,curr}$ is outside the registration region, based on which the model of the retina curvature has been established, the auto-focus algorithm will only use the gradient-ascent approach given in \autoref{alg:image_feedback}.

If the CR score is above the optimal image quality threshold $T_2$, the desired axial motion $\Delta X_{des,a}$ is set to zero and the probe stops moving. Otherwise, a score below $T_2$ indicates that the optimal image quality has not been reached and further adjustment is needed. The algorithm will check if the position inferred by the prior model has been achieved, i.e. $F_M$ is \textit{TRUE} or not, and if the user has moved laterally to the tissue surface by comparing the lateral motion $\|K_c\Delta X_{probe} \|$ with \textit{ROBOT RESOLUTION}. If the model-inferred position has been reached while the user has not moved laterally, i.e. if $F_M$ is \textit{TRUE} and $\|K_c\Delta X_{probe} \| <$ \textit{ROBOT RESOLUTION}, the prior model differs from the actual retina curvature due to external factors, e.g., patient movement or registration error; in this case, \autoref{alg:image_feedback} will be applied to fine-tune the image quality using gradient-ascent image optimizer. In any other case, the probe will move to the axial position specified by the prior-model and the flag is set accordingly, i.e. $\Delta X_a$ is set to the difference between model inferred position $M(K_cX_{probe,curr})$ and $K_aX_{probe,curr}$. If it is found that $\Delta X_a$ is less than \textit{ROBOT RESOLUTION}, the position inferred from prior model has been reached and flag $F_M$ is set to TRUE. Otherwise, if $\Delta X_a$ is larger than \textit{ROBOT RESOLUTION}, the position inferred from the prior model has not yet been reached and $F_M$ is set to FALSE.

\begin{algorithm}[h]
\SetAlgoLined
\KwIn{Image ($I$), Prior Model ($M$), Current/Previous probe position ($X_{probe,curr}, X_{probe,prev})$}
\KwOut{Desired Movement ($\Delta X_{des,a}$)}
 $q_I = CR(I)$\;
 $\Delta X_{probe} = X_{probe,curr} - X_{probe,prev}$ \; 
 $F_M$=TRUE\;
 \uIf{$X_{probe,curr}$ outside registration region}{
  $\Delta X_{des,a} =$ \autoref{alg:image_feedback} ($I$)\;
 }
 \Else{
    \uIf{$q_I \geq T_2$}{
      $\Delta X_{des,a} = 0$ \;
     }
    \uElseIf{$F_M$ \& $\| K_c\Delta X_{probe} \| < $ ROBOT RESOLUTION}{
        $\Delta X_{des,a} =$ \autoref{alg:image_feedback} ($I$)\;
    }
    \Else{
        $\Delta X_{des,a} = M(K_cX_{probe,curr}) - K_aX_{probe,curr}$\;
        \uIf{$\Delta X_{des,a} <$ ROBOT RESOLUTION}{
            $F_M=$TRUE\;
        }
        \Else{
            $F_M=$FALSE\;
        }
    }
 }
 \caption{Axial Control Algorithm, with Prior Model}
 \label{alg:model_image}
\end{algorithm}

\subsection{\textbf{Mid-Level Optimizer and Low-Level Controller}}
\label{sssec:coop-midlow}
The desired motion ${\dot{X}}_{des}$, which is the output of the high-level controller, is then sent to the mid-level optimizer, which calculates the equivalent desired joint values $q$ while satisfying the limits of the robot's joints. The mid-level optimizer is formulated as 

\begin{align}
    \begin{split}
        \underset{\dot{q}_{des}}{min} | {J\dot{q}} &- {\dot{X}}_{des} | \\
        \dot{q}_{L} \le {\dot{q}} \leq {\dot{q}}_{U} ~ &\& ~ q_L \leq q \leq q_U
    \end{split}
\end{align}
where $J$ is the Jacobian of the SHER. ${q}_L$, ${q}_U$, ${\dot{q}}_{U}$, and ${\dot{q}}_{U}$ denote the lower and upper limits of the joint values and velocities, respectively.

The desired joint values are then sent to the low-level PID controller of the SHER, as a result of which the desired objectives generated based on the high-level controller are satisfied.

\section{Hybrid Teleoperated Framework}
\label{ssec:teleoperated}
In the previous section, the proposed hybrid strategy was explained using the hybrid cooperative framework. In this section, the hybrid teleoperated framework will be discussed. In addition to filtering tremors of surgeon's hand, the hybrid teleoperated framework enables large scaling down of the motions to make precise and minute manipulations inside the confined areas of the eyeball. The proposed hybrid teleoperated framework has the same patient-side platform, mid-level optimizer and low-level controller architecture at the SHER side as the hybrid cooperative platform. However, the lateral motion of the pCLE probe is controlled remotely by the surgeon.

\begin{figure*}[thpb]
    \centering
    \includegraphics[width=\linewidth]{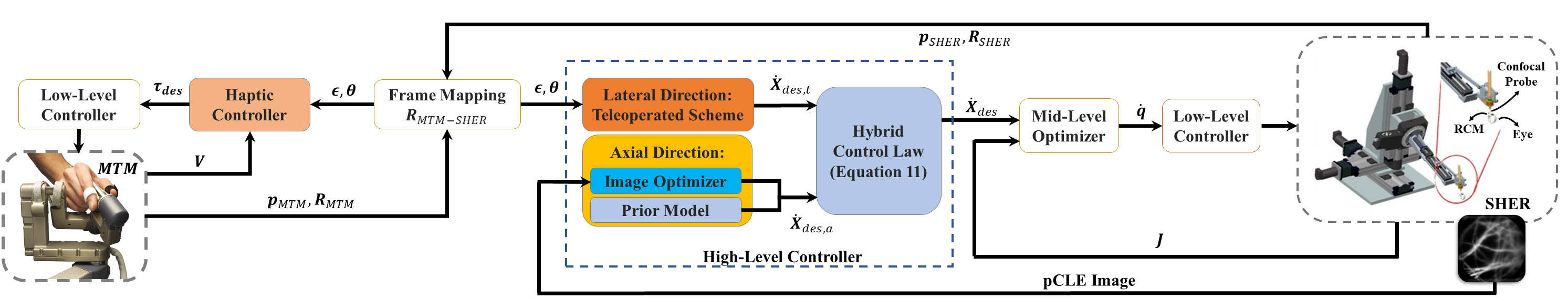}
    \caption{General schematic of the proposed hybrid teleoperated control strategy. Hybrid Control Law is described in \autoref{eq:hybrid_motion_tele}. Lateral Direction control is described in \autoref{sssec:tele-lateral}. Axial Direction control is described in \autoref{sssec:coop-axial-grad} and \autoref{sssec:coop-axial-model}. Mid-Level Optimizer and Low-Level Controller are described in \autoref{sssec:coop-midlow}. Haptic Controller is described in \autoref{sssec:tele-haptic}.}
    \label{fig:teleop_block_diagram}
\end{figure*}

For this purpose, the dVRK system has been used to enable remote control of the SHER. The dVRK system is an open-source telerobotic research platform developed at Johns Hopkins University. The system consists of the first-generation da Vinci surgical system with a master console, including the surgeon interface with stereo display and MTM to control the pCLE probe. 

The proposed hybrid teleoperated framework differ from the previous hybrid cooperative framework in three aspects: 1) a hybrid control law using the teleoperated commands (\autoref{sssec:tele-hybrid-law}), 2) a teleoperated scheme to control the lateral direction (\autoref{sssec:tele-lateral}), and 3) a haptic controller on the MTM side (\autoref{sssec:tele-haptic}). \autoref{fig:teleop_block_diagram} shows the block diagram of the proposed hybrid teleoperated framework.

\subsection{High-Level Controller}
\subsubsection{\textbf{Hybrid Control Law}}
\label{sssec:tele-hybrid-law}
Like the hybrid cooperative framework, the hybrid motion controller of the teleoperated platform has the lateral and axial components at the SHER side with the following formulations:
\begin{equation}
    \label{eq:hybrid_motion_tele}
    \dot{X}_{des} = {K}_t \dot{X}_{des,t} + {K}_a \dot{X}_{des,a}
\end{equation}
where subscripts $t$ and $a$ refer to the teleoperated (the lateral) and autonomous (the axial) components of the motion, respectively; $K_t$ is the motion specification matrix along the lateral direction same in the hybrid cooperative framework, i.e., $K_t=K_c$. Also, $\dot{X}_{des,a}$ is the desired axial motion component of the SHER, which is derived using the autonomous control strategy discussed in \autoref{sssec:coop-axial-grad} and \autoref{sssec:coop-axial-model} to maximize the image quality. 

\subsubsection{\textbf{Lateral Direction - Teleoperated Scheme}}
\label{sssec:tele-lateral}
The desired lateral component of the motion, $\dot{X}_{des,t}$, is received from a remote master console as: 
\begin{align}
    {\dot{X}}_{des,t} &= \begin{bmatrix}
        {\dot{p}}_{des,t} \\
        {\dot{R}}_{des,t} 
    \end{bmatrix}
\end{align}
such that
\begin{align}
    \begin{split}
        {\dot{p}}_{des,t} &= \frac{{\epsilon}}{\Delta t} \\
        {\dot{R}}_{des,t} &= \frac{{\theta}}{\Delta t}
    \end{split}
\end{align}
where all the above variables are expressed in the base frame of the SHER; ${\dot{p}}_{des,t}$ and ${\dot{R}}_{des,t}$ refer to the translational and rotational components of the desired user-commanded velocities. ${\epsilon}$ and ${\theta}$ denote the current translational and rotational components of the user-commanded increments, which are calculated as the \textit{tracking error} of the SHER. The \textit{tracking error}, \textit{i.e.} the difference between the commanded position (Cartesian position of the MTM transformed into the base frame of the SHER) and the actual position of the pCLE probe, can be written as
\begin{align}
    \begin{split}
        {\epsilon} &= \beta \cdot {R}_{MTM-SHER}^{-1} \cdot {\Delta p}_{MTM} - {\Delta p}_{SHER} \\
        {\theta} &= Rodriguez \left[ {R}_{SHER}^{-1} \cdot {R}_{MTM-SHER}^{-1} \cdot {R}_{MTM} \right]
    \end{split}
    \label{eqn:tracking_error}
\end{align}
where $\beta$ is the teleoperation motion scaling factor;
${R}_{MTM-SHER}$ is the orientation mapping between the base frames of the MTM and SHER; ${R}_{MTM}$ and ${R}_{SHER}$, respectively, denote the rotation of the MTM in its own base frame and the rotation of pCLE probe in the base frame of the SHER. ${\Delta p}_{MTM}$ refers to the translation offset between the current position of the MTM (${p}_{MTM}$) and the initial position of the MTM (${p}_{MTM,0}$) in the base frame of MTM. ${\Delta p}_{SHER}$ denotes the translation offset between the current position (${p}_{SHER}$) and the initial position (${p}_{SHER,0}$) of the pCLE probe in the base frame of the SHER.
\begin{align}
    \begin{split}
        {\Delta p}_{MTM} &= {p}_{MTM} - {p}_{MTM,0}\\
        {\Delta p}_{SHER} &= {p}_{SHER} - {p}_{SHER,0}
    \end{split}
\end{align} 
This relative position and absolute orientation control setup allows the surgeon to control the pCLE probe intuitively.

The output of the high-level hybrid controller given in \autoref{eq:hybrid_motion_tele}, is then sent to the mid-level optimizer and low-level PID controller at the SHER side (discussed in \autoref{sssec:coop-midlow}) to fulfill the desired motion.
\subsection{\textbf{Haptic Controller at the MTM Side}}
\label{sssec:tele-haptic} 
To provide the surgeon with sensory situational awareness from the SHER side, a haptic controller is designed at the MTM side. 

The first component of an effective haptic feedback strategy is a gravity compensation to cancel out the weight of the MTM. To enable a zero-gravity low-friction controller, a multi-step least square estimation/compensation approach is used, which also addresses elastic force and friction modeling in the dynamics model. Details of the gravity compensation algorithm can be found in the \href{https://github.com/jhu-dvrk/dvrk-gravity-compensation}{\textit{open-source code}.}

Besides, the haptic controller implemented at the MTM side includes a  compliance wrench defined in the MTM base frame, to enable haptic feedback. The compliance wrench has two parts:
\begin{itemize}
    \item an \textit{elastic} part, which is proportional to the tracking error (\autoref{eqn:tracking_error}) of the SHER, and
    \item a \textit{damping} part, which is proportional to the Cartesian linear velocities of the MTM
\end{itemize}
The compliance wrench $W$ can be formulated as
\begin{align}
    \begin{split}
        W & = \begin{bmatrix}
        F_p \\
        F_R 
        \end{bmatrix}\\
        F_p &= k_p \epsilon + b V\\
        F_R &= k_R \theta
    \end{split}
\end{align}
where $k_p$ and $k_R$ are the elasticity coefficient of translation error $\epsilon$ and rotation error $\theta$, respectively; $b$ is the damping coefficient of  $V$, the Cartesian linear velocities of the MTM (defined in the base frame of the MTM). The compliance wrench is then converted to the corresponding joint torque $\tau_W$ using the inverse Jacobian $J^{-1}$ as
\begin{align}
    \tau_W = J^{-1} W
\end{align}
The final desired torque is the summation of the compliance wrench torque $\tau_W$ and gravity compensation torque $\tau_{gc}$, as
\begin{align}
    \tau_{des} = \tau_W + \tau_{gc}
\end{align}
The desired torque is then sent to the low-level PID controller of the MTM to achieve the desired haptics feedback.

\section{Experiment Results}
\label{sec:experiment}
\autoref{fig:Experiment_Setup} shows the experiment setup. Elbow support was provided to the user to interact with the SHER. The sampling frequency of the pCLE is set to be 60 Hz. The mid-level optimizer frequencies of the SHER and dVRK are both 200 Hz. The cooperative gain, $\alpha$, is set to be 10 \micro m/s per $1~N$. The motion scaling factor of the teleoperated framework, $\beta$, is set to be 15 \micro m/s per 1 mm/s. The elasticity and damping coefficients of the haptics feedback are set to $50~N/m$ and $5~\frac{N}{m/s}$, respectively. The software framework is built upon the \textit{CSA} library developed at Johns Hopkins University \cite{Chalasani2018}. The eyeball phantom used was built in-house with spherical shape of 30 mm outer diameter. A thin layer of tissue paper with an uneven surface is attached to the inside of the eyeball phantom, representing the detached retina. A 10 mm opening is made at the top to simulate the open-sky process. The phantom is 3D printed with ABS material. Before the experiment, the tissue was stained with 0.01\% acriflavine. We chose acriflavine because of its ready availability. For clinical applications, fluorescein, which is bio-compatible, should be used.

To validate the effectiveness of the proposed frameworks, three experiments were conducted, as discussed below. In this set of experiments, a user highly familiar with the SHER and the dVRK was chosen to eliminate any extra variability over the experiment outcome due to the insufficient skill level of the user. 
\subsection{Experiment \#1}
\begin{figure}[h]
    \centering
    \subfloat{\includegraphics[width=0.12\linewidth]{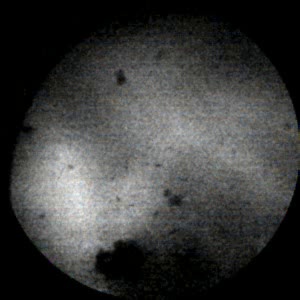}}
    \hspace{0.5pt}
    \subfloat{\includegraphics[width=0.12\linewidth]{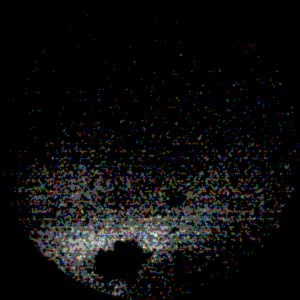}}
    \hspace{0.5pt}
    \subfloat{\includegraphics[width=0.12\linewidth]{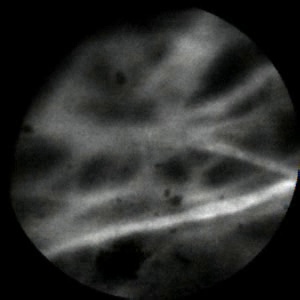}}
    \hspace{0.5pt}
    \setcounter{subfigure}{0}
    \subfloat[]{\includegraphics[width=0.12\linewidth]{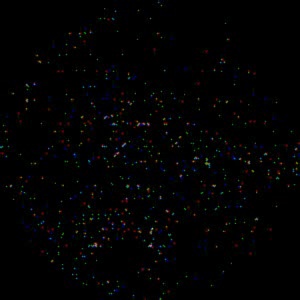}\label{fig:manual_scan}}
    \hspace{0.5pt}
    \subfloat{\includegraphics[width=0.12\linewidth]{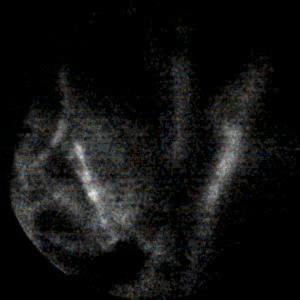}}
    \hspace{0.5pt}
    \subfloat{\includegraphics[width=0.12\linewidth]{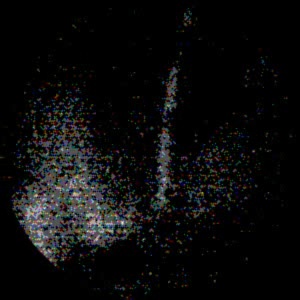}}
    \hspace{0.5pt}
    \subfloat{\includegraphics[width=0.12\linewidth]{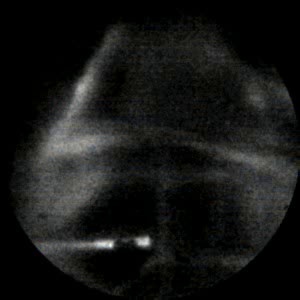}}
    
    \subfloat{\includegraphics[width=0.12\linewidth]{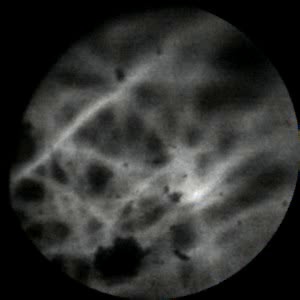}}
    \hspace{0.5pt}
    \subfloat{\includegraphics[width=0.12\linewidth]{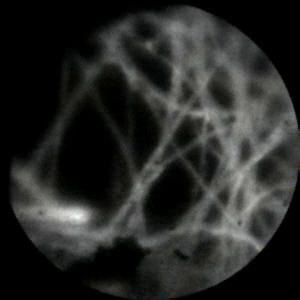}}
    \hspace{0.5pt}
    \subfloat{\includegraphics[width=0.12\linewidth]{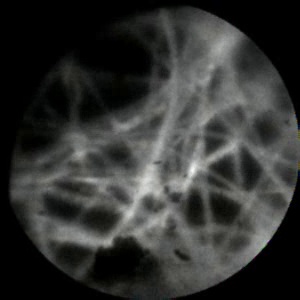}}
    \hspace{0.5pt}
    \setcounter{subfigure}{1}
    \subfloat[]{\includegraphics[width=0.12\linewidth]{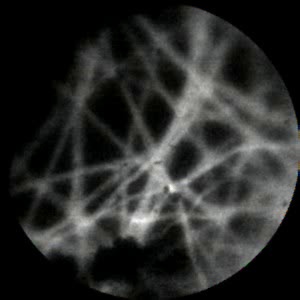}\label{fig:hybrid_teleop_scan}}
    \hspace{0.5pt}
    \subfloat{\includegraphics[width=0.12\linewidth]{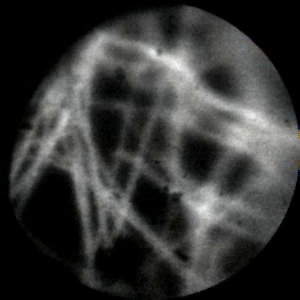}}
    \hspace{0.5pt}
    \subfloat{\includegraphics[width=0.12\linewidth]{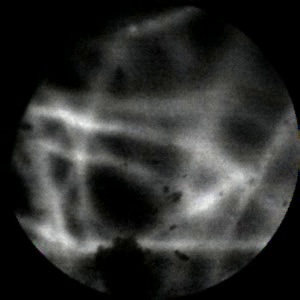}}
    \hspace{0.5pt}
    \subfloat{\includegraphics[width=0.12\linewidth]{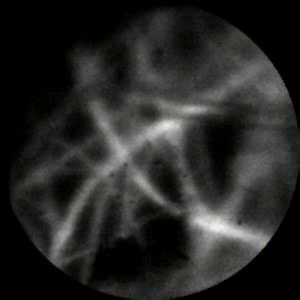}}
    
    \caption{A sequence of images acquired from (a) manual scanning, (b) the proposed hybrid teleoperated framework, sub-sampled at 1.5 Hz. As can be seen, the manual scanning image quality is far from desirable.}
\end{figure}

\begin{figure}[h]
    \centering
    \includegraphics[width=0.35\linewidth]{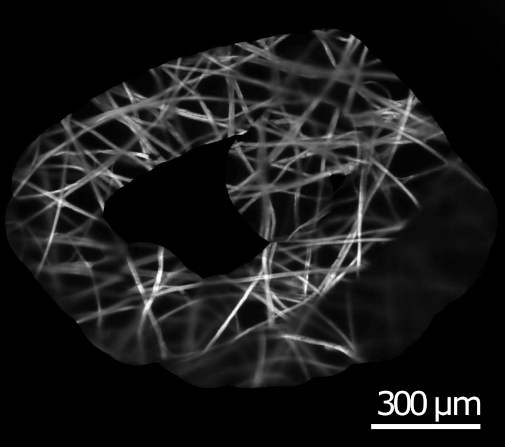}
    \caption{An example of generated mosaicking images.}
    \label{fig:mosaicking}
\end{figure}

The purpose of this experiment was to demonstrate the impracticality of manual pCLE scanning. As discussed previously, manual control of a pCLE probe within the micron-order focus range is extremely challenging, if not impossible, due to several factors including hand tremor and patient motion. In this experiment, the user was instructed to try his/her best to control the probe for a clear scanning stream of images. The user, however, was unsuccessful in performing the task. \autoref{fig:manual_scan} shows a sequence of images acquired during this task. As can be seen, the quality of the images is far from desirable. By way of comparison, a sequence of images acquired using the proposed hybrid teleoperated framework is also presented in \autoref{fig:hybrid_teleop_scan}, which shows a considerable improvement of the image quality, and an example of the generated mosaicking image is shown in \autoref{fig:mosaicking}. The complete result of this experiment can be found in the video supplemental material of the paper. 

\vspace{-0.5cm}

\subsection{Experiment \#2}
\begin{figure}[h]
    \centering
    \subfloat[]{\includegraphics[width=0.48\linewidth]{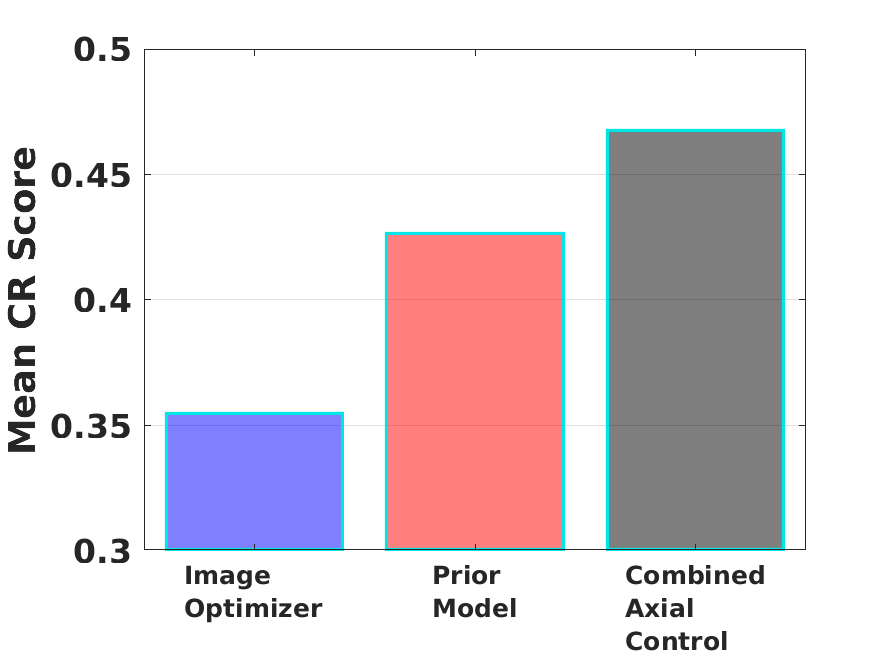}\label{fig:experiment_cr}}
    \hspace{1pt}
    \subfloat[]{\includegraphics[width=0.48\linewidth]{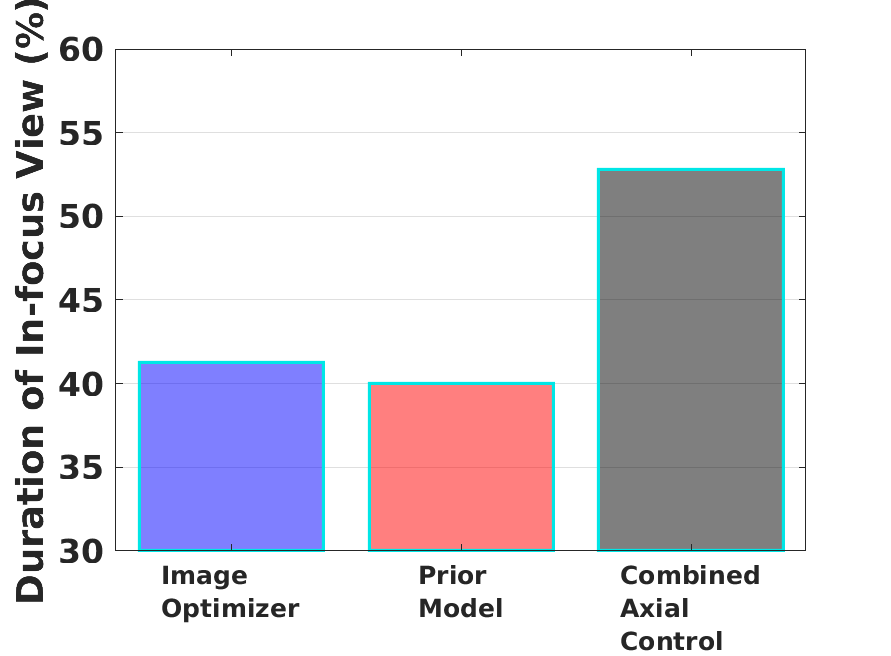}\label{fig:experiment_infocus_duration}}
    \captionsetup{width=0.8\linewidth}
    \caption{Experiment results of the image optimizer only, prior model only, and combined axial controller. Comparison of (a) CR scores, (b) duration of in-focus view.}
\end{figure}

The second experiment was designed to validate the improvement due to the addition of the prior model to the image optimizer as previously proposed in \cite{li2019novel}. For this purpose, three experiments were conducted with: 1) only the gradient-ascent image optimizer (no prior model), 2) only the prior model (no image optimizer), and 3) the combined axial controller (image optimizer and prior model) as proposed in this work. In these experiments, the task was defined to follow a triangular path with a side length of approximately 3 cm using the teleoperated setup. Two metrics were used to evaluate and compare the efficacy of the three experiments:
\begin{enumerate}
    \item \textbf{Mean CR score}: The average of CR scores throughout the scanning task, whose formulation was given in \autoref{sssec:coop-axial-grad}. 
    \item \textbf{Duration of In-focus View}: The percentage of the instances throughout the scanning task that the pCLE probe is in-focus (indicated as a CR score higher than the threshold $T_2$).
    \end{enumerate}
\autoref{fig:experiment_cr} and \autoref{fig:experiment_infocus_duration} present a comparison of the two metrics for the three experiment scenarios. The combined axial controller outperformed both the image optimizer only and the prior model only approaches. The mean CR score of the combined controller is $0.47$, higher than that of the prior model only approach $0.43$, and the image optimizer only approach $0.35$. The in-focus duration of the combined controller was $53\%$, also higher than that of the prior model only approach $40\%$, and the image optimizer only approach $41\%$. A higher mean CR score and longer in-duration focus make the combined axial controller more effective than using only either of the individual components.
\subsection{Experiment \#3}
The third experiment was designed to compare the performance with and without autonomous axial control using the cooperative and teleoperated frameworks. Same as Experiment \#2, the task was defined to follow a triangular path with a side length of approximately 3 cm. The user was instructed to try his/her best to maintain the optimal image quality during the task. In this set of experiments, the orientation of the pCLE probe was locked for more accurate trajectory comparison and easier maneuvering for the user.

\begin{figure}[!thpb]
    \centering
    \subfloat[]{\includegraphics[width=0.48\linewidth]{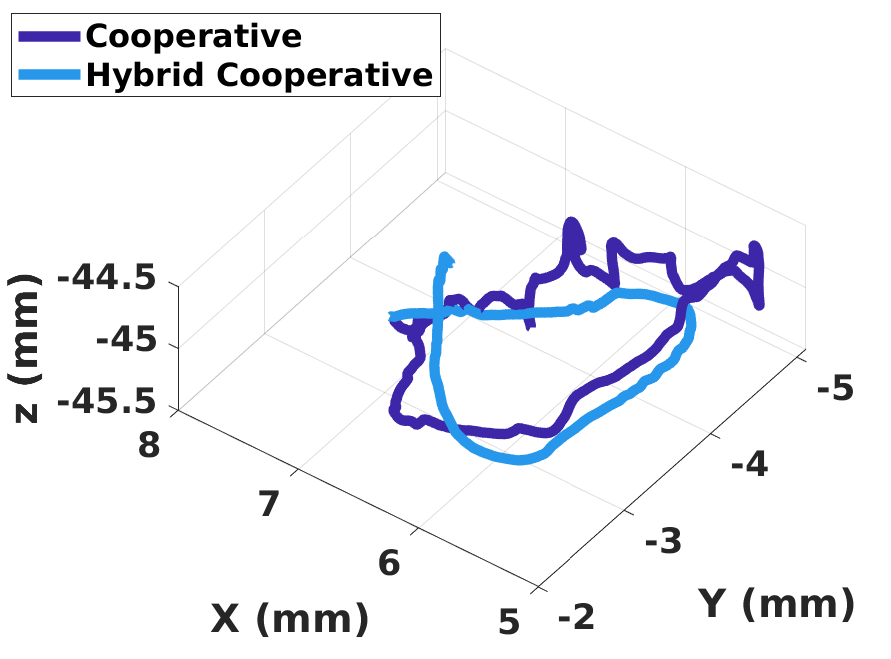}\label{fig:experiment_path_adm}}
    \hspace{1pt}
    \subfloat[]{\includegraphics[width=0.48\linewidth]{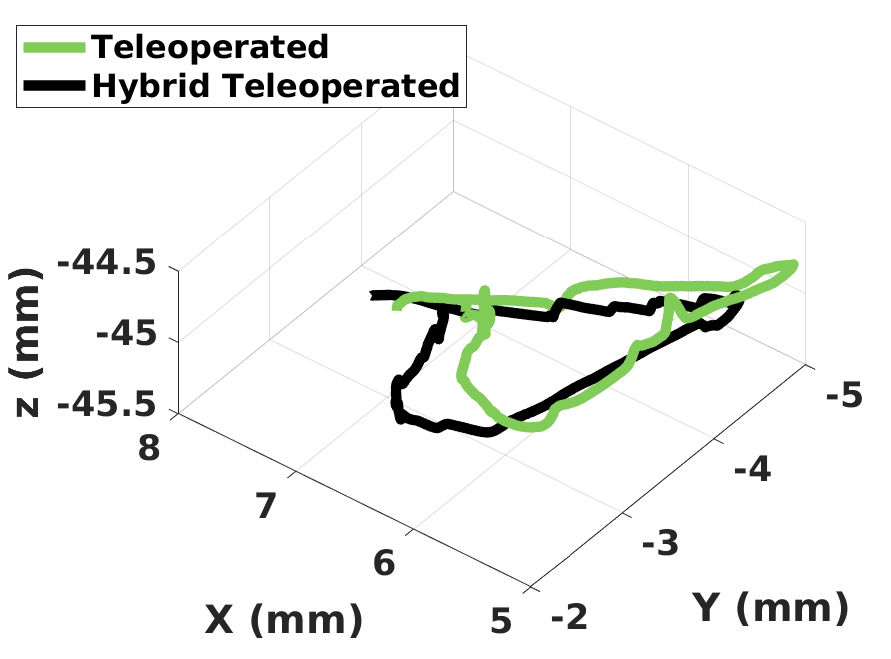}\label{fig:experiment_path_tele}}
    \captionsetup{width=0.8\linewidth}
    \caption{Comparison of scanning path between (a) Cooperative and hybrid cooperative frameworks, (b) Teleoperated and hybrid teleoperated frameworks.}
\end{figure}

\autoref{fig:experiment_path_adm} and \autoref{fig:experiment_path_tele} show a comparison of the 3-D path between the frameworks (the proposed hybrid cooperative vs. traditional cooperative, and the proposed hybrid teleoperated vs conventional teleoperated). As shown in \autoref{fig:experiment_path_adm}, the proposed hybrid cooperative framework resulted in a significantly smoother path with lower elevational variability along the axial direction ($Z$-axis), as compared to the traditional cooperative approach. Smaller deviations along the axial direction of the probe shaft are an indication of better (less jerky) manipulation and better control of the probe around the optimal probe-to-tissue distance.

Interestingly, as can be seen in \autoref{fig:experiment_path_tele}, the teleoperated and hybrid teleoperated frameworks resulted in a \textit{visually} comparable elevational motion along the axial direction. Both trajectories appear to be smooth \textit{visually}. To further assess the motion made during these two cases, Motion smoothness (MS) \cite{shahbazi2018multimodal} was calculated as a quantitative measure of smoothness. MS is calculated as the time-integrated squared jerk, where jerk is defined as the third-order derivative of position:
\begin{equation}
    MS = \frac{1}{N}\sum_{i=0}^{N} \sqrt{J_{x,i}^2 + J_{y,i}^2 + J_{z,i} ^2}
\end{equation}
where
\begin{align*}
    J_{u,i} = \frac{u_{i+\frac{3}{2}}-3u_{i+\frac{1}{2}}+3u_{i-\frac{1}{2}}-u_{i-\frac{3}{2}}}{\Delta t^3} 
\end{align*}
is the third-order central finite difference of the discrete trajectory signal along $u$-axis. A smoother trajectory will result in a smaller jerk, thus smaller MS value. Based on the experiment, the MS scores are 6.477E${-2}$ for the cooperative framework, 3.481E${-2}$ for the hybrid cooperative framework, 4.201E${-2}$ for the teleoperated framework, and 4.202E${-2}$ for the hybrid teleoperated framework. The hybrid cooperative framework outperformed the traditional cooperative framework by $46.3\%$, while the two teleoperated frameworks yield almost the same motion smoothness scores. The quantitative evaluation of MS yields the same result as the visual inspection. The total duration of the scanning task was 61.92 s for the cooperative framework, 35.95 s for the hybrid cooperative framework, 60.66 s for the teleoperated framework, and 42.43 s for the hybrid teleoperated framework. 

\section{User Study}
\label{sec:userstudy}
\subsection{Study Protocol}
To further evaluate and compare the proposed frameworks, a set of user studies was conducted. The study protocol was approved by the Homewood Institutional Review Board, Johns Hopkins University. In this study, 14 participants without clinical experience were recruited at the Laboratory for Computational Sensing and Robotics (LCSR), with all participants being right-handed.

The participants were asked to perform a path-following task in four cases:
\begin{itemize}
    \item using the traditional cooperative framework, \textit{i.e.}, without the auto-focus algorithm
    \item using the proposed hybrid shared cooperative framework, \textit{i.e.}, with the auto-focus algorithm
    \item using the traditional teleoperated framework, \textit{i.e.}, without the auto-focus algorithm
    \item using the proposed hybrid shared teleoperated framework, \textit{i.e.}, with the auto-focus algorithm
\end{itemize}

The task was defined to perform pCLE scanning of a triangular region of a side length of approximately 3 mm within an eyeball phantom using the four setups. The participants were instructed to perform the scanning task while trying their best to maintain the best image quality. \autoref{fig:user_view} shows the view provided to the users during the experiments.

Before each trial, the participants were given around 10 minutes to familiarize themselves with the system before they proceed with the main experiments. The order of the experiments was randomized to eliminate the learning-curve effect. After each trial, the participants were asked to fill out a post-study questionnaire. The questionnaire included a form of the NASA Task Load Index (NASA TLX) survey for evaluation of operator workload. Data recording started when the participants pressed the activation pedal of the robot. This ensured consistent and trackable start timing between participants.

In our previous study \cite{li2019novel}, we observed that manipulating the robot at micron-level precision within the confined space of the eye is very challenging for novices. Therefore, the orientation of the pCLE probe was locked to reduce the complexity of the scanning task.

\subsection{Metrics Extraction and Evaluation}
To evaluate task performance during four experiments, 6 quantitative metrics were used: CR score, duration of in-focus view, and MS (discussed above); as well as task completion time, Cumulative Probability for Blur Detection (CPBD), and Marziliano Blurring Metric (MBM). A NASA TLX questionnaire including 6 qualitative metrics answered by the participants was also studied.  

While CR, MBM, and CPBD are image-quality metrics, they use different approaches to assess the sharpness of an image. These metrics were chosen to validate the consistent improvement of the image quality regardless of the metric type, demonstrating the generalizability and consistency of the outcome.

\subsection{Results and Discussion}
\begin{figure}
    \centering
    \includegraphics[width=0.75\linewidth]{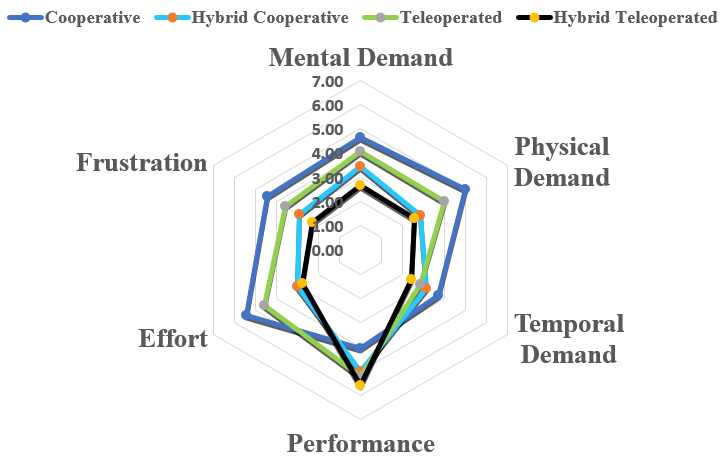}
    \caption{NASA TLX questionnaire result}
    \label{fig:nasa_tlx}
\end{figure}

\autoref{fig:nasa_tlx} shows the results of the NASA TLX questionnaire for 14 users. The questionnaire includes six criteria: mental demand, physical demand, temporal demand, performance level perceived by the users themselves, level of effort, and frustration level, each with a maximum value of 7. Single-factor ANOVA analysis was used to statistically evaluate the results, and statistical significance was observed in the mental demand (p-value=3.0E${-2}$), physical demand (p-value=2.0E${-3}$), effort (p-value=6.3E${-5}$) and frustration level (p-value=1.0E${-2}$), while no statistical significance was observed in the temporal demand (p-value=0.182) and performance (p-value=0.062). 

A Tukey's Honest Significance test was followed for the four categories with statistical significance (w.s.s). The mental demand has decreased from $5.43$ for the cooperative framework to $2.78$ for the hybrid teleoperated framework. The physical demand has decreased from $5.00$ for the cooperative framework and $4.00$ for the teleoperated framework, to $2.57$ for the hybrid teleoperated framework. Similarly, the effort level has decreased from $5.43$ for the cooperative framework and $4.57$ for the teleoperated framework, to $2.78$ for the hybrid teleoperated framework. A lower frustration level was observed in the hybrid teleoperated framework 2.29 compared to the cooperative framework 4.43. Out of the 14 users, 11 indicated that the hybrid teleoperated framework is the most preferred modality. 

\begin{figure}[thpb]
    \centering
    \subfloat[]{\includegraphics[width=0.45\linewidth]{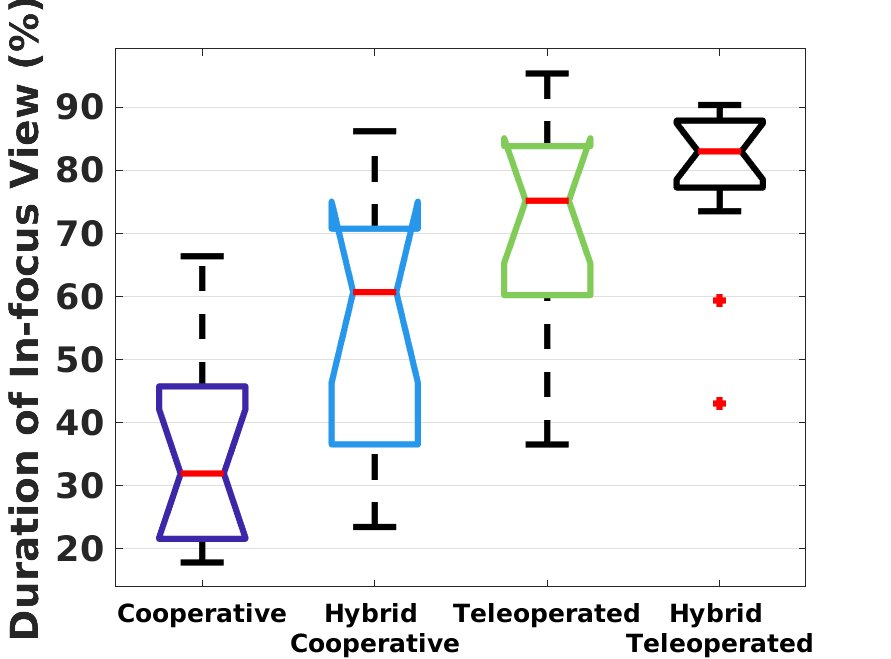}\label{fig:infocus_duration}}   
    \hspace{1pt}
    \subfloat[]{\includegraphics[width=0.45\linewidth]{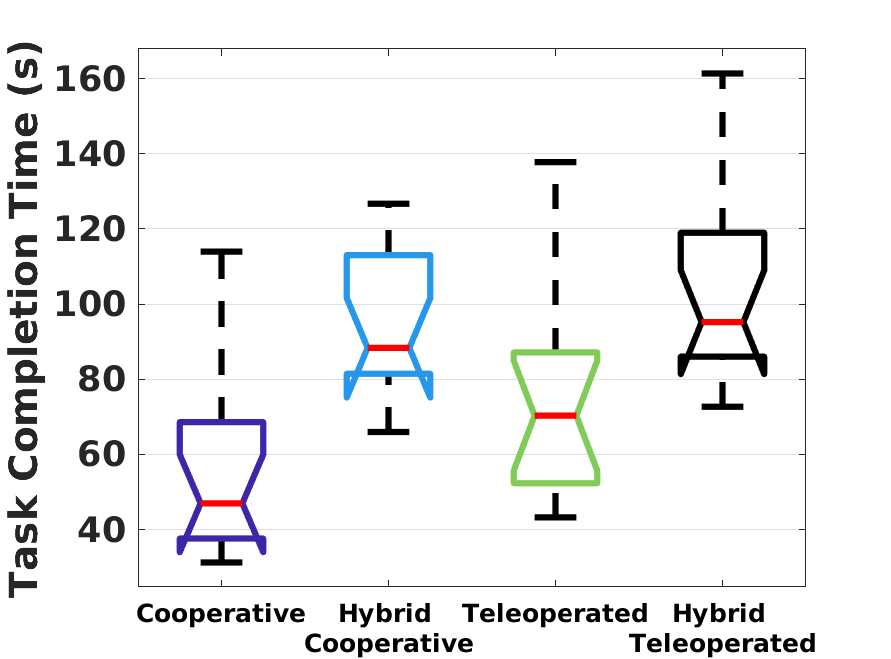}\label{fig:total_duration}}
    
    \subfloat[]{\includegraphics[width=0.43\linewidth]{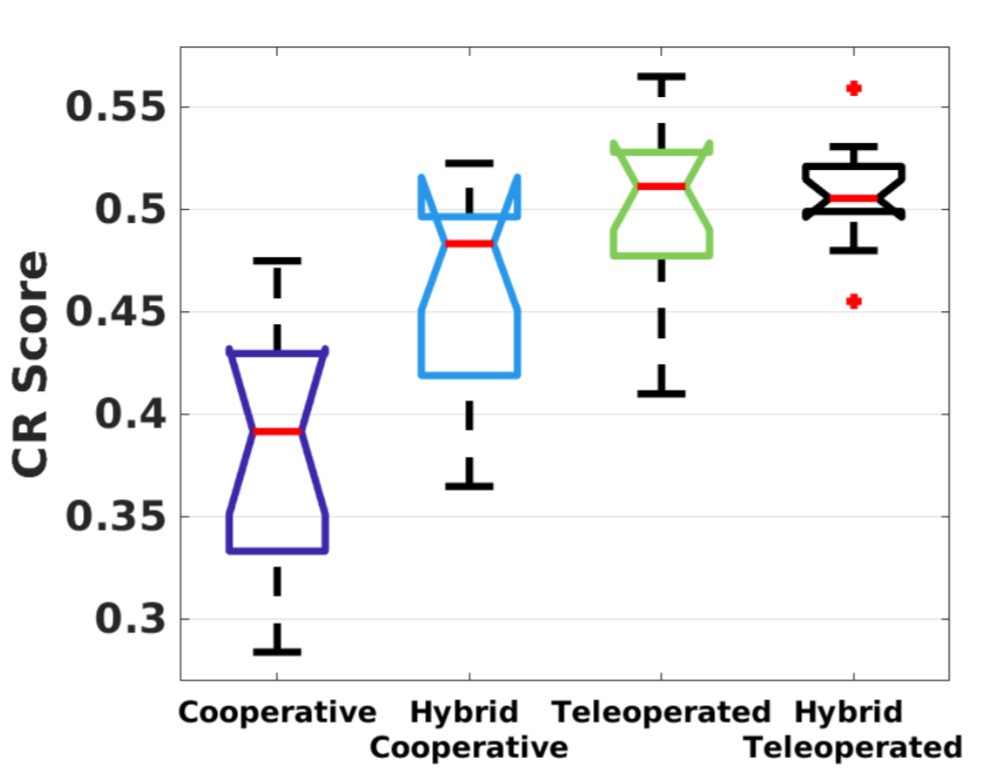}\label{fig:cr_score}}
    \hspace{1pt}
    \subfloat[]{\includegraphics[width=0.45\linewidth]{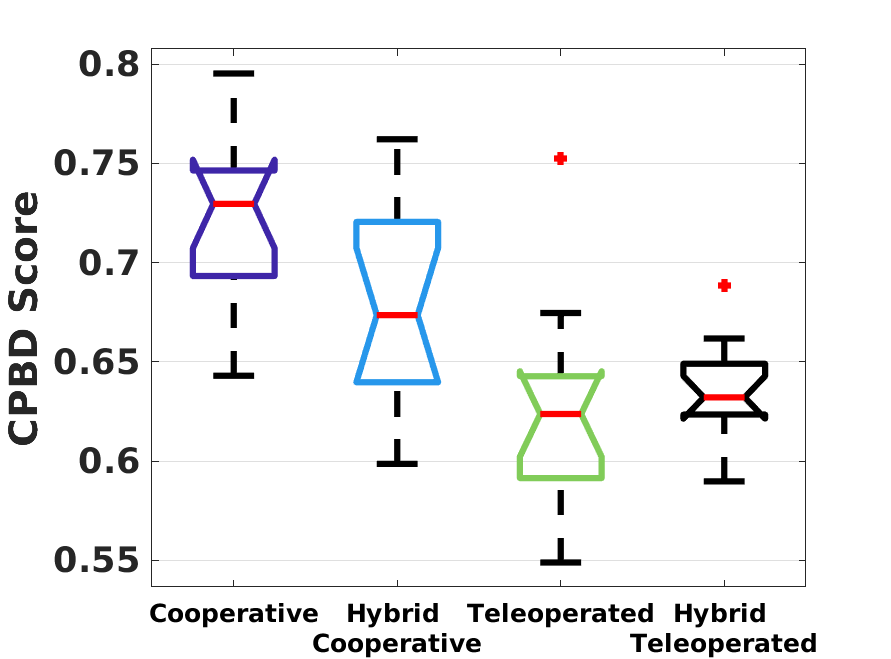}\label{fig:cpbd_score}}
    
    \subfloat[]{\includegraphics[width=0.45\linewidth]{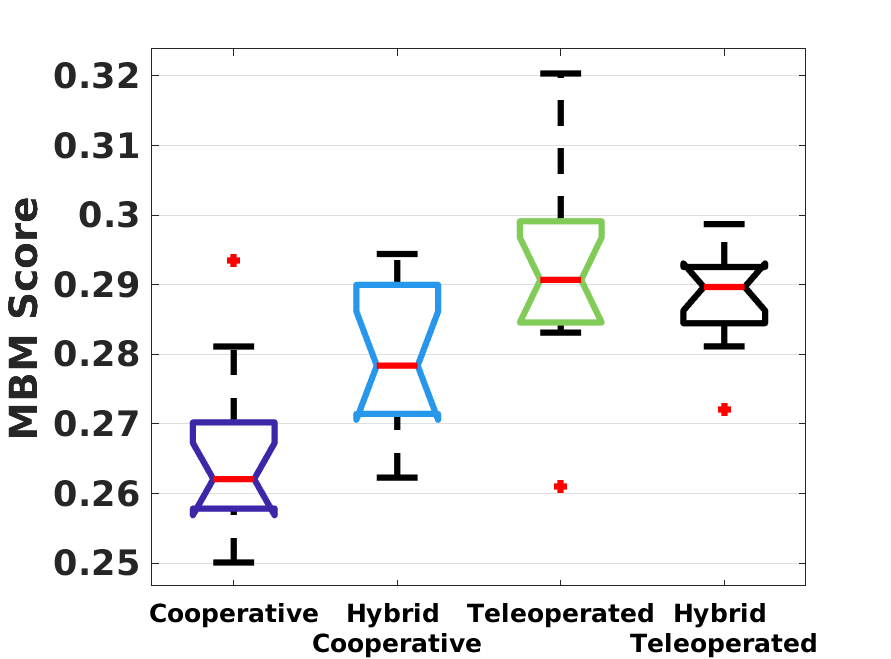}\label{fig:marz_score}}
    \hspace{1pt}
    \subfloat[]{\includegraphics[width=0.45\linewidth]{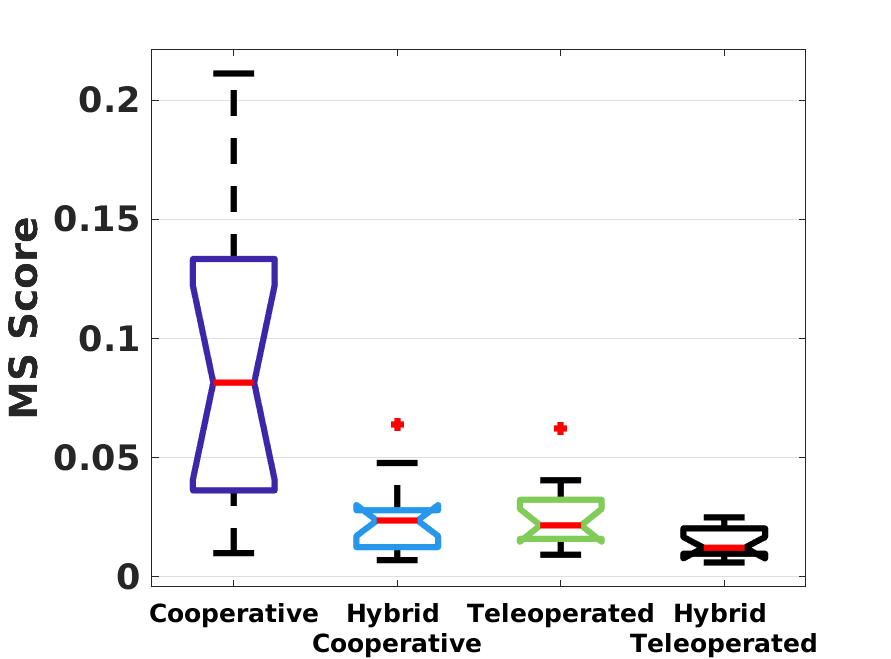}\label{fig:ms_score}}
    \caption{Results of the user study: (a) Duration of in-focus view, (b) Task completion time, (c) CR score, (d) CPBD score, (e) MBM score, and (f) MS score.}
    \label{fig:experiment_result}
\end{figure}

The quantitative results are shown as boxplots in \autoref{fig:experiment_result} with six metrics compared: in-focus duration, task completion time, CR, CPBD, MBM and MS. Applying the Single-factor ANOVA analysis, statistically significant differences were observed between various modes of operation for all the metrics, with \textit{p-values} equal to 1.0E${-8}$, 4.7E${-5}$, 1.1E${-9}$, 4.4E${-7}$, 2.7E${-7}$, and 9.2E${-8}$ respectively. 

\textit{Post hoc} analysis using the Tukey's test showed that in terms of in-focus duration, both the hybrid teleoperated ($79\%$) and teleoperated ($73\%$) frameworks outperformed the hybrid cooperative ($56\%$) and cooperative ($36\%$) frameworks w.s.s. The hybrid cooperative framework also outperformed the cooperative framework w.s.s. The mean CR scores of the hybrid teleoperated (CR=0.51), teleoperated (CR=0.50) and hybrid cooperative (CR=0.46) frameworks were better than that of the cooperative framework (CR=0.38) w.s.s. The mean MBM score of the teleoperated framework (MBM=0.2920) was better than the hybrid cooperative (MBM=0.2792) and cooperative  (MBM=0.2657) frameworks w.s.s. The hybrid teleoperated (MBM=0.2886) and hybrid cooperative frameworks also outperformed the cooperative framework w.s.s.

Task completion time, CPBD, and MS (metrics with a lower value indicating a higher performance) were also analyzed with Tukey's test as discussed below. The task completion time for the hybrid teleoperated framework (105s) was longer than the teleoperated (76s) and cooperative frameworks (58s) w.s.s. The hybrid cooperative framework (95s) also took longer than the cooperative framework (58s) w.s.s. However, it should be noted that the task completion time for the hybrid teleoperated framework and hybrid cooperative framework include pre-operative eyeball registration time as well. Extracting the pre-operative registration time, the task completion time for the hybrid teleoperated framework and hybrid cooperative framework become 51s and 44s, respectively.

The CPBD score for the hybrid teleoperated framework (CPBD=0.6342) was lower than the cooperative framework (CPBD=0.7233) w.s.s. The CPBD score for the teleoperated framework (CPBD=0.6270) was lower than that of both hybrid cooperative (CPBD=0.6773) and cooperative frameworks w.s.s. The hybrid cooperative framework also had lower CPBD compared to the cooperative framework w.s.s. The MS scores for the hybrid teleoperated (MS=1.407E{-2}), teleoperated (MS=2.608E{-2}) and hybrid cooperative (MS=2.514E{-2}) frameworks were smaller than that for the cooperative framework (MS=8.944E{-2}) w.s.s.

In general, the hybrid cooperative framework was shown to be more advantageous over the traditional cooperative framework both qualitatively and quantitatively. The user workload decreased when using the hybrid cooperative framework, and the image quality was considerably improved. The comparison between the hybrid teleoperated framework and the traditional teleoperated framework showed that the hybrid teleoperated framework reduced the user's workload while providing an equally high-quality image. 

Comparing the hybrid teleoperated framework proposed in this work with the hybrid cooperative framework previously proposed in \cite{li2019novel}, the hybrid teleoperated framework demonstrated better performance with a higher percentage of duration where the pCLE probe is in-focus. A reason for the improved performance can be the ability for a large motion scaling, which is unavailable in a cooperative setting. Overall, the hybrid teleoperated system demonstrated clear advantages: quantitatively consistent high-quality images and qualitatively with $78.6\%$ of the users indicating it as the most favorable framework.

However, several limitations may be addressed in future studies. First, none of users were clinicians, but with a wide range of skills in controlling robotic frameworks. A part of our future work will focus on comparing and evaluating the four frameworks with a larger set of users including surgeons. Secondly, in this study, an artificial eyeball phantom was used and technical aspects of the work were demonstrated and validated. In the future, cadaveric eyes will be used along with image mosaicking and clinical studies to evaluate the systems in clinical settings. Lastly, the stain material should be switched to fluorescein for bio-compatibility.

\section{Summary}
\label{sec:conclusion}
A novel hybrid strategy was proposed for real-time endomicroscopy imaging for retinal surgery. The proposed strategy was deployed on two control frameworks - cooperative and teleoperated. The setup consists of the dVRK, SHER, and a distal-focus pCLE system. The hybrid frameworks allow surgeons to scan the area of interest in the retina without worrying about the loss of image quality and hand tremor. The effectiveness of the hybrid frameworks was demonstrated via a series of experiments in comparison with the traditional teleoperated and cooperative frameworks. A user study of 14 users showed that both hybrid frameworks lead to statistically significant lower workload and improved image quality both qualitatively and quantitatively.

\bibliographystyle{IEEEtran}
\bibliography{IEEEabrv,ref}

\end{document}